\PassOptionsToPackage{dvipsnames,table,xcdraw}{xcolor}
\documentclass{article} % For LaTeX2e
\usepackage{iclr2026_conference,times}

\usepackage{amsmath,amsfonts,bm}

\def\eqref#1{equation~\ref{#1}}
\def\1{\bm{1}}

\DeclareMathAlphabet{\mathsfit}{\encodingdefault}{\sfdefault}{m}{sl}
\SetMathAlphabet{\mathsfit}{bold}{\encodingdefault}{\sfdefault}{bx}{n}

\usepackage{xcolor}
\usepackage{hyperref}
\hypersetup{
    colorlinks=true,
    linkcolor=orange,
    filecolor=magenta,
    urlcolor=orange,
    citecolor=orange,
    pdftitle={VIA: Visual Interface Agent for Robot Control},
    pdfauthor={Hengyuan Hu, Jensen Gao, Priya Sundaresan, Satvik Sharma, Jeannette Bohg, Dorsa Sadigh},
}
\usepackage[capitalise, nameinlink]{cleveref}
\usepackage{url}
\usepackage{graphicx}
\usepackage{wrapfig}
\usepackage{booktabs}
\usepackage{capt-of}  % \captionof, so a table can share a float with a figure
\usepackage{array}
\usepackage[normalem]{ulem}  % strike through old review text without changing emphasis
\usepackage{amssymb}  % bent arrow in the control-setup table
\usepackage[most]{tcolorbox}  % shaded, page-breakable cards for the tool docs appendix
\usepackage{multirow}
\usepackage{enumitem}
\usepackage[varqu,scaled=0.95]{inconsolata}  % nicer monospace font for code (varqu = straight code quotes)
\usepackage[T1]{fontenc}  % real underscore glyphs in \texttt (OT1 fakes \_ with a rule)

\title{\fontsize{16}{19}\selectfont VIA: Visual Interface Agent for Robot Control}

\author{Hengyuan Hu, Jensen Gao, Priya Sundaresan, Satvik Sharma, \\
\textbf{Jeannette Bohg, Dorsa Sadigh} \\
Stanford University}

\newcommand{\via}{\textsc{VIA}}

\definecolor{aiNew}{HTML}{173F7A}
\definecolor{aiOld}{HTML}{808080}
\DeclareRobustCommand{\ai}[2]{%
  {\color{aiNew}\hypersetup{citecolor=aiNew,linkcolor=aiNew,urlcolor=aiNew}#1}%
  \if\relax\detokenize{#2}\relax\else
    \space{\color{aiOld}\hypersetup{citecolor=aiOld,linkcolor=aiOld,urlcolor=aiOld}\sout{#2}}%
  \fi}

\iclrfinalcopy % Show the authors and omit review line numbers.
\begin{document}

\maketitle
\fancyhead{} % The ICLR style sets its conference header inside \maketitle.
\renewcommand{\headrulewidth}{0pt}

% Author correspondence.
\makeatletter
\begin{NoHyper}
{\long\def\@makefntext#1{\noindent #1}\footnotetext{Correspondence to hengyuan.hhu@gmail.com}}
\end{NoHyper}
\makeatother

% Provenance tags in section files, per paragraph:
%   [adapted: <file>]  = taken from website text with light rewording
%   [expanded: ...]    = built on website content plus new material
%   [free]             = written fresh for the paper
\begin{abstract}
Robot manipulation is a complex task that requires visual perception, physical reasoning, planning, and closed-loop control.
General-purpose foundation models (FMs) have grown remarkably capable of some of these, especially perception and reasoning.
Inspired by the growing ability of FM-powered agents to operate software through visual interfaces, we ask whether that same competence suffices to control a robot \emph{directly} given the right interface.
We present \via{} (\textbf{V}isual \textbf{I}nterface \textbf{A}gent), a framework that recasts robot control as an agentic task where an off-the-shelf agent drives a manipulator directly through an interface.
The interface is a virtual workspace of interactable visual components where the agent can probe pixels of interest with mouse-like tools and command a virtual end-effector to set new target poses with a small set of general tools.
We show that \via{} enables agents of varying capabilities to perform diverse tabletop manipulation tasks in simulation. We also use \via{} to drive an off-the-shelf mobile manipulation robot in the real world to complete tasks that require both navigation and manipulation.
Both settings are zero-shot, requiring no robot training or additional action primitives.
Performance scales well with the size and strength of the underlying FMs.
These results suggest that frontier agents already possess skills that transfer directly to robot control given the right interface: \emph{your coding or computer-use agent is, in a sense, secretly a robot-control agent}.
\end{abstract}

\section{Introduction}
\label{sec:intro}

General robot manipulation is a long-standing research goal at the intersection of robotics and learning.
It requires policies that are capable of complex visual understanding, physical reasoning, long-term planning, and precise closed-loop control.
Foundation models (FMs)~\citep{radford2019language, bommasani2021foundation, openai2023gpt4} trained on internet-scale data have become remarkably capable of several of these, such as visual perception and reasoning, which has made leveraging them for robotic manipulation highly attractive.

One approach for this is to convert existing vision-language models into vision-language-action (VLA) models, by fine-tuning them on robot data to produce low-level actions~\citep{brohan2023rt2, black2024pi0}.
However, due to a lack of both compute and robot-specific data, VLAs are often orders of magnitude smaller than frontier FMs.
This gap has a profound impact on capabilities, particularly the physical reasoning and long-horizon planning skills that current robot policies lack.

Another paradigm is Code-as-Policies (CaP)~\citep{liang2023code, singh2023progprompt}, where FMs generate programs to control robots using perception APIs and hand-crafted skill primitives, e.g., \texttt{mask = segment\_text\_prompt("green cube")}, \texttt{grasp\_pose = sample\_grasp\_pose("red cube")}~\citep{fu2026capx}.
Although this is a natural way to tap into the strong coding capabilities of FMs, the FMs are not directly involved during rollouts.
Therefore, the system is often bottlenecked by human-crafted perception and action abstractions rather than by the FM.

\begin{figure}[t]
\centering
\includegraphics[width=1.0\textwidth]{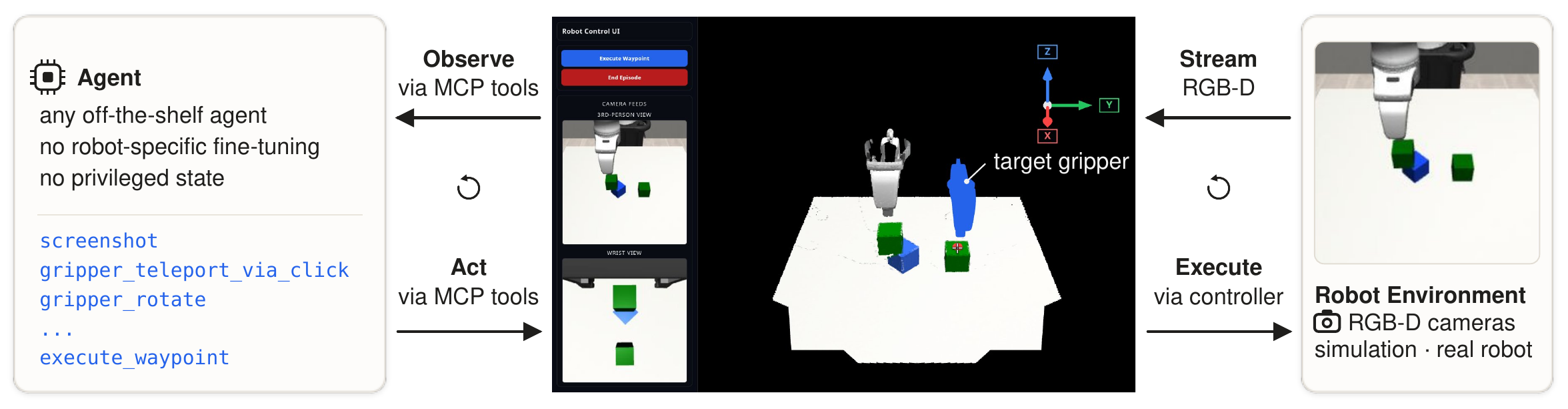}
\caption{\textbf{\via~overview:} an off-the-shelf agent controls a robot through an \emph{interactive} interface.
\textbf{Middle:} The interface includes a target gripper (blue) rendered on top of the real scene to form a virtual workspace.
In this example, the workspace contains a 3D point-cloud scene reconstructed from RGB-D cameras, as well as raw third-person and wrist camera feeds.
In real-world scenarios where the point clouds are sparse and the depth measurements are noisy, we render the virtual gripper on top of raw RGB camera feeds.
\textbf{Left:} The agent observes and acts through human-legible MCP tools: e.g., it can take screenshots, move the virtual gripper with \texttt{gripper\_\{ops\}}, observe the resulting change in the interface (\emph{without moving the actual gripper}), and refine the target through additional tool calls.
\textbf{Right:} Once satisfied with the target gripper's state (the \emph{waypoint}), the agent invokes a controller to move the real gripper to that target.
The agent repeats this loop, replanning from a new observation after each tool call and completing the task through a sequence of waypoints.}
\label{fig:front}
% Keep the gap below this figure compact.
\vspace{-10pt}
\end{figure}

We propose a new way to connect FMs to robot manipulation: recasting robot control as a visual tool-use task for agents by introducing a virtual workspace that the agent can probe and interact with, similar to general computer use.
Agents~\citep{yao2023react} wrap FMs with harnesses to enable better performance on long-horizon tasks~\citep{yang2024sweagent, jimenez2024swebench}, as well as closed-loop control of software or games in virtual environments~\citep{anthropic2025claude37}.
Frontier agents can now operate real software through native screenshots, clicking, and keyboard operations~\citep{shi2017worldofbits, anthropic2024computeruse, openai2025operator}.
They perform long-horizon tasks in a closed-loop manner, iteratively observing, planning, and executing, remarkably similar to what we want from robot policies.
Therefore, we ask: given the right \emph{visual interface} and human-legible tools, %akin to mouse and keyboard
can a state-of-the-art agent control a robot without any robot-specific adaptation?

We present \via~(\textbf{V}isual \textbf{I}nterface \textbf{A}gent), a framework for robot control that lets off-the-shelf visual agents drive robot manipulators through an interface (\cref{fig:front}) and a controller.
The interface is a virtual workspace where the agent interacts with the scene, staging and refining controller commands without affecting the real environment.
The core component is a virtual \emph{target gripper} that the agent can move around with a small set of human-legible Model Context Protocol (MCP) tools~\citep{anthropic2024mcp}, much as a human would pose an object in 3D design software such as Blender.
Through the interface, the agent reasons about the scene, issues MCP tool calls, and inspects their visual effects to refine the target gripper's state until satisfied.
The agent submits the refined target gripper state to the controller as a \emph{waypoint}, and the controller moves the real gripper to that target.
The agent then observes the updated scene and plans the next waypoint, repeating until the task is complete.

We evaluate \via~with two frontier agents, Claude Code~\citep{claudecode2025} and Codex~\citep{openai2025codex}, across various model tiers.
Our evaluation suite includes six simulated tabletop manipulation tasks and five real-world mobile manipulation tasks, spanning pick-and-place, articulated objects, precise execution, and long-horizon multi-object assembly.
\via~achieves strong results, solving these tasks end-to-end in a zero-shot setting with high success rates at an estimated API cost of a few dollars per successful episode on most tasks.
We show that \via's performance scales well with the underlying model capacity, achieving success rates of $38\%$, $73\%$, and $100\%$ with Codex using Terra 5.6, Sol 5.6, and Astra 6, respectively, and showing a similar trend for Claude Code.
\via~can also consume demonstrations written as plain text: providing a list of example waypoints increases \via's success rate with Claude Code from $62\%$ to $76\%$ and with Codex from $68\%$ to $74\%$ on simulated tasks.
\via~opens up a new opportunity for leveraging state-of-the-art FMs at scale in robotics and demonstrates that, with the right interface, robot control can become another economically valuable agentic task that benefits directly from modern models and agents.
Since \via~uses frontier models and agents unmodified, improvements in agent perception, reasoning, context management, and tool use transfer directly to robot control.

\section{Related Work}
\label{sec:related}

We review the following areas related to our work: existing ways to leverage FMs for robot manipulation, waypoint-based robot control, and computer-use agents.

\subsection{Foundation Models for Robot Manipulation}
% There are four main categories of approaches to leverage FMs for robot manipulation.

\smallskip \textbf{Vision-Language-Action (VLA) models.}
VLAs fine-tune vision-language models (VLMs) on robot data to directly predict low-level actions~\citep{brohan2023rt2, kim2024openvla, black2024pi0}.
While these models have shown that FMs can enhance robot policies, their generalization abilities often still lag behind those of their base FMs, in part because fine-tuning on action data can degrade their reasoning and multimodal understanding abilities~\citep{hancock2025vlm2vla}.
% Furthermore, compute constraints typically restrict VLAs from leveraging the largest and most powerful FMs available.
% \via~uses off-the-shelf agents for robot control directly, avoiding fine-tuning that may degrade generalization.

\smallskip \textbf{FMs in Hierarchical Robot Control.}
Hierarchical approaches use FMs for high-level planning and reasoning while delegating low-level control to learned robot policies.
SayCan~\citep{ahn2022saycan} combines language-model planning with learned or scripted skill affordances to select which skill to execute.
Text2Motion~\citep{lin2023text2motion} extends this by verifying the geometric feasibility of the planned skill sequence with learned skill value functions.
More recent approaches use VLMs to guide VLAs with subtask instructions~\citep{shi2025hirobot, chen2026volo}, 3D manipulation policies with predicted paths~\citep{li2025hamster}, or diffusion policies with latent plans~\citep{huang2025thinkact}.
FMs can also guide policy learning rather than execution: \citet{debakker2025scaffolding} use VLM-generated keypoint trajectories as scaffolds and dense rewards to train a residual RL policy in simulation, so the VLM is no longer needed at test time.

% These systems delegate motion generation to learned policies or action primitives; \cref{sec:setup} contrasts this with ARC.}{}

\smallskip \textbf{Visual Prompting for Manipulation.}
PIVOT~\citep{nasiriany2024pivot} queries a VLM to select among visually annotated action candidates sampled from an iteratively refined Gaussian proposal distribution.
MOKA~\citep{liu2024moka} similarly prompts a VLM to select affordance keypoints and coarse waypoint parameters from marked images, relying on object segmentation, grasp proposals, and hand-designed motion routines to generate robot motion.
% \via~lets the agent observe raw camera images and directly specify controller targets (6-DoF end-effector poses, 3-DoF mobile-base poses, and binary gripper commands) without visual and action-proposal primitives.

\smallskip \textbf{Code-as-Policies.}
Code-as-Policies (CaP) uses FMs to write code to control a robot~\citep{liang2023code, singh2023progprompt}.
The generated code, rather than the FMs themselves, is the policy that handles perception and control.
As a result, the code policies often rely on perception APIs like SAM-based segmentation and CLIP-based object retrieval to parse image observations, and rely on action primitives like grasping and placement modules~\citep{liang2023code, huang2023instruct2act, huang2023voxposer} to handle actions.
CaP-X~\citep{fu2026capx} adds multi-turn execution feedback and allows the coding FMs to revise their generations after each attempt, but the policy still does not leverage the general power of the FMs during execution.

\smallskip \textbf{Comparison with \via.}
\via~operates in a distinct setting different from the four categories listed above.
Compared to VLAs, \via~uses off-the-shelf FMs and agents without any fine-tuning.
In contrast to VLMs in hierarchical settings, \via~does not involve any low-level VLAs for translating instructions into robot actions.
\via~allows agents to specify controller targets directly in the controller's native degrees of freedom (DoF), avoiding any action abstractions or action-proposal primitives used in visual prompting.
Finally, unlike CaP, which uses the generated program as the policy, the agent in \via~\emph{is} the policy and remains in the control loop during execution.

% %  to interface with perception and control modules.
% % \ai{}{By avoiding task-specific training, CaP directly leverages the general reasoning and perception abilities of constituent FMs, often resulting in better generalization than end-to-end approaches that require robot data.}
% \ai{Instruct2Act~\citep{huang2023instruct2act} places SAM and CLIP behind perception APIs, while VoxPoser~\citep{huang2023voxposer} composes perception-grounded 3D value maps in code for motion planning.
% CaP-X~\citep{fu2026capx} adds multi-turn execution feedback, visual differencing, skill synthesis, and ensembled reasoning.}{}
% \ai{}{However, code may not always be the optimal interface for FMs to control robots.
% For example, CaP depends on a predefined library of perception and control primitives (e.g., object detectors, pose estimators, or hand-crafted skills).}
% CaP can be highly sensitive to the design of these primitives~\citep{fu2026capx}, thus requiring careful, potentially task-specific engineering.

% \ai{}{Like CaP, \via~also provides FMs with an engineered interface for robot control, but its interface is task-agnostic and exposes only generic kinematic control and raw sensor views.}
% In the taxonomy of \citet{fu2026capx}, \via~operates below their lowest abstraction tier: the agent commands generic 6-DoF waypoints and receives no perception primitives.
% \ai{}{Perception and reasoning are deferred to the FM agent itself when possible, allowing for direct scaling with FM capabilities.}

\subsection{Waypoint-based Robot Control}

Waypoint-based policies shorten the decision horizon by predicting sparse end-effector targets executed by a controller or motion planner~\citep{shridhar2023peract, goyal2023rvt}.
HYDRA~\citep{belkhale2023hydra} and SPHINX~\citep{sundaresan2025sphinx} combine waypoints for long-range motion with dense actions for precise manipulation.
\via~builds on SPHINX's visual teleoperation interface.
Instead of having a human teleoperator collect data using this interface, \via~uses off-the-shelf FM agents to specify and refine waypoints to control the robot directly without robot-specific training.

\subsection{Computer-Use Agents}

A rapidly growing line of work equips FMs to operate computer software directly through visual interfaces and a small set of generic tools (e.g., taking screenshots, clicking, typing)~\citep{anthropic2024computeruse, openai2025operator}.
Benchmarks such as OSWorld~\citep{xie2024osworld} and WebArena~\citep{zhou2024webarena} measure this competence on real desktop and web environments, and coding agents extend it to multi-step software tasks~\citep{jimenez2024swebench, claudecode2025}.
These agents exhibit strong closed-loop visual agency: they perceive an interface, act through tools, observe the outcome, and recover from mistakes.
\via~repurposes this competence for robot manipulation by simply presenting the robot as another application the agent can operate, so that progress in computer-use agents transfers directly to robot control.

\section{VIA: Visual Interface Agent for Robot Control}
\label{sec:method}

% core idea
\via~recasts robot control as an agentic observe-act loop over a visual interface.
% overall loop, observe, act, feedback
An agent operates a robot-control UI much as a person would: it takes a screenshot, reasons about what it sees, and issues intuitive commands through Model Context Protocol (MCP) tools that stage a virtual target gripper visually representing the input to the controller.
It then observes the result and adjusts until satisfied before invoking the controller to actually move the robot.
It continues the loop until the task is complete or the episode times out.
We describe the interface in \cref{sec:method-interface} and the agentic loop in \cref{sec:method-agent}.
The detailed list of MCP tools is provided in \cref{sec:appendix-tools}.

% % most important detail observe
% The agent observes only what the interface renders and has no access to privileged simulator state.
% For simplicity, we also do not provide any perception-assisting APIs such as segmentation functions, so visual understanding is solely the job of the agent and its underlying model.
% % most important detail on act
% The MCP tools are low-level operations that set the target pose and gripper state of the end-effector, e.g., moving the gripper via \texttt{gripper\_drag}, \texttt{gripper\_translate}, or \texttt{gripper\_rotate}, or changing its open/close state via \texttt{gripper\_toggle}.
% We also provide a few functions to operate the camera and to show extra visual helpers on the UI.

% most important detail for the agent

\subsection{Visual Interface}
\label{sec:method-interface}

\begin{figure}[t]
\centering
\includegraphics[width=\textwidth]{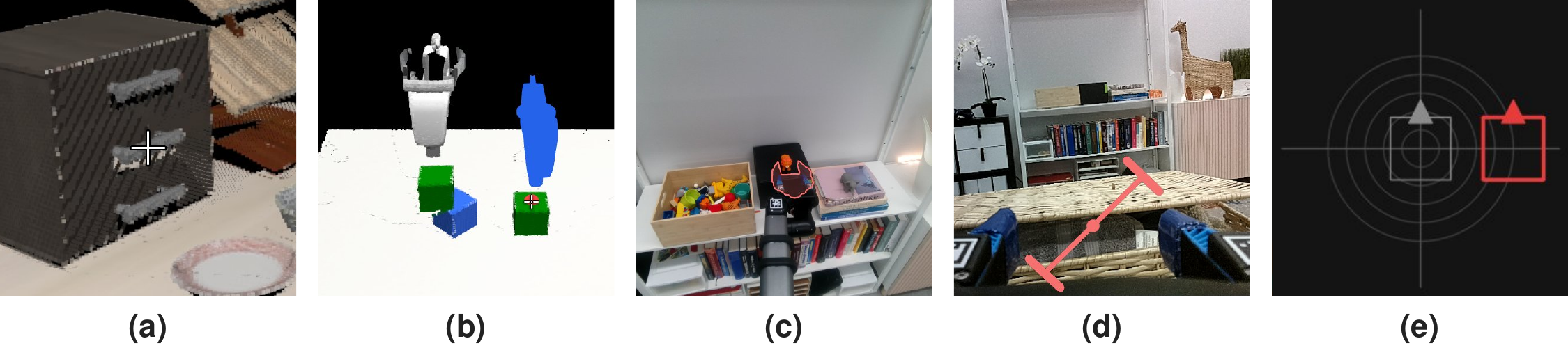}
\caption{\textbf{Examples of interactions in \via.}
The core idea of \via~is to provide visual feedback after each tool call so that the agent can experiment with the target in this virtual workspace.
(a) \textit{hover}: draws a crosshair at snapshot coordinates $(u, v)$ so the agent can check whether its predicted point falls on the intended target (e.g., the handle of the middle drawer).
It also returns depth information when available.
(b) \textit{click}: moves the target gripper (blue) to an approach position near the red point clicked by the agent.
(c) \textit{translate}: moves the target gripper (red, contour) given 3-DoF offsets.
(d) \textit{rotate}: rotates the target gripper (red, glyph) given 3-DoF rotation deltas.
(e) \textit{nav}: shows the staged movement of the 3-DoF mobile base, when available.
}
\label{fig:interface-interactions}
\end{figure}

The interface is a collection of interactive visual components that the agent can observe from and interact with.
We first explain how the agents observe the world through the interface, and then how they control the robots by interacting with it.

All interface components are constructed from raw sensor inputs and robot proprioceptive information, without access to privileged state.
When the depth readings are dense and reliable, such as in tabletop settings or in simulation, we reconstruct a 3D point cloud from RGB-D inputs and use it as the main workspace.
When the depth readings are sparse, such as in our real-world experiments where a mobile base operates in a room, we simply show the 2D images.
The 3D point cloud allows the user (either a human or an AI agent) to change the viewport by applying the \texttt{orbit}, \texttt{pan}, or \texttt{zoom} functions to a virtual camera, enabling views beyond those of the fixed cameras used to construct the point cloud.
The agent observes this point cloud through snapshots of the current viewport, which it can modify, rather than the raw vectors of the points.
% This workspace also resembles 3D design software (e.g., Blender) and 3D games that the model may have been trained on, so its 3D reasoning capabilities may transfer.
The interface can be presented as a graphical user interface with a defined layout (as shown in the center of \cref{fig:front}), or passed directly to the agent as a structured list of named, typed entries, such as JPEG image bytes and proprioceptive values.

% We also put two raw camera feeds on the left: a third-person view for understanding the overall scene, and a wrist view for close-up analysis during fine manipulation.
% Note that an agent observes the UI by taking screenshots of the whole UI, including the camera feeds, and it never reads the raw vectors of points in the point cloud.

% control
An agent controls the robot by deciding inputs to the controller.
For example, when controlling a single robot arm with a gripper, the controller inputs are 6-DoF target poses and a binary open/close state; when controlling an arm with a mobile base, the controller may additionally take 3-DoF base movement parameters.
We refer to this set of target values as a \emph{waypoint}.
Instead of imagining and setting a precise target value directly given RGB-D inputs, the interface in \via~acts as a virtual workspace that allows the agent to probe and experiment with the target visually.

\cref{fig:interface-interactions} illustrates the most representative interactions that the interface supports.
The core component is a virtual illustration of the \emph{target waypoint} that visually shows the staged target transformation.
For a robot arm manipulator, the target gripper is rendered as a 3D model in the point-cloud workspace or as a contour or line-and-dot glyph over a camera image.
For a mobile manipulator, this may also include a visualization of the base movement.
Each adjustment updates the visualization, allowing the agent to inspect and revise the target without moving the real robot.
Additionally, helper tools such as \texttt{hover} render visual cues to help the agent verify its prediction or probe additional information from the environment, such as depth, when available.

% To control the robot in this interface, users set the 6-DoF target pose of the end-effector and a binary open/close state by manipulating a blue \emph{target gripper} via mouse and keyboard operations.
% Users can click in the point cloud to teleport the target gripper near the clicked point, drag it with the mouse, or use the keyboard to translate it by exact metric offsets.

% Similarly, users can rotate the target gripper via the mouse by dragging it along a gizmo axis, or via the keyboard by typing a rotation axis and angle, e.g., \texttt{\{x -45 enter\}}.

%  loop

% credits, and hint that it will also work in real
% This interface and waypoint-style control are largely borrowed from the teleoperation interface of SPHINX~\citep{sundaresan2025sphinx}, which demonstrated that humans can operate a similar interface to easily collect robot training data in both simulation and the real world.

\begin{figure}[t]
\centering
\includegraphics[width=1.0\textwidth]{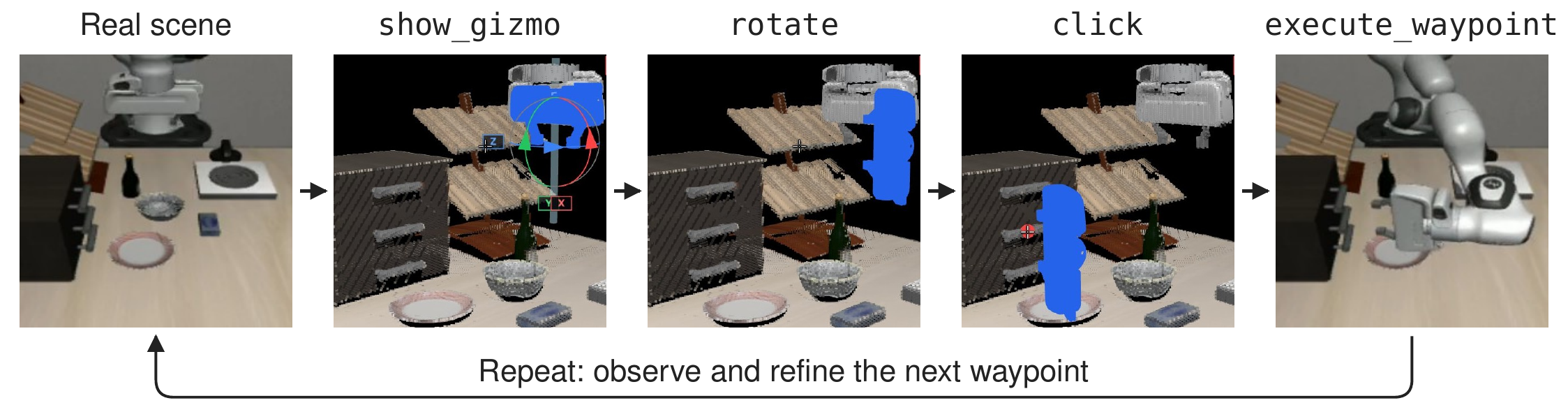}
\caption{\textbf{One round of the \via~control loop}, shown on the \emph{open drawer} task.
From left to right, the agent observes the real scene, then refines a waypoint through \texttt{show\_gizmo}, \texttt{rotate}, and \texttt{click} (tool names abbreviated).
The three target previews show successive states of the same waypoint, with each tool call returning an updated snapshot.
Once satisfied with the waypoint, the agent calls \texttt{execute\_waypoint} to pass the waypoint to the controller, which moves the real robot.
Then the agent observes the outcome and plans the next waypoint, repeating until termination.
Note that the viewport has been adjusted by the agent to look towards the drawer in previous tool calls.}
\label{fig:loop}
% Keep the gap below this figure compact.
\vspace{-10pt}
\end{figure}

\subsection{Agentic Loop}
\label{sec:method-agent}

%  proposed rewrite
An agent controlling a robot through a controller runs a simple loop: observe the scene, call the controller with a waypoint, observe the outcome, and repeat until a terminal condition is met.
This closed-loop structure lets the agent recover from errors and re-plan based on what it observes, but the only feedback on a waypoint arrives after the robot has already moved, so every misjudged target costs a real motion.

With the interface, \via~keeps this outer loop but nests a \emph{mini-loop} inside each iteration.
Instead of committing a waypoint in one shot, the agent edits the target waypoint through any number of interface operations, each of which returns an updated rendering of the staged target on top of the real scene, and keeps revising until satisfied.
Then the agent calls \texttt{execute\_waypoint}, which invokes a controller that moves the real gripper to the target pose while obeying physical constraints.
The mini-loop allows for trial-and-error to happen in the virtual workspace rather than on the physical robot: the agent verifies each waypoint visually before executing it, so the real robot only ever receives waypoints that have already been checked virtually.
\cref{fig:loop} shows one round of this loop, with \via~refining and then executing a single waypoint to prepare the gripper for opening a drawer.

%  overview
% With the MCP tools, any agent that takes images as input and is compatible with MCP can control the robot to perform useful tasks.
% We use Claude Code (CC)~\citep{claudecode2025} and Codex~\citep{openai2025codex} with their latest models to drive the system.
% We choose these agents not for writing code, as we are not following the Code-as-Policies paradigm, but simply because they are the most widely adopted, easy-to-use agents wrapping the most powerful models.
% Our goal is to show that general agents not specifically built for robot control already possess skills such as general perception, spatial reasoning, and planning that make them capable robot policies when equipped with the right interface.
% This setup also opens up new opportunities for leveraging state-of-the-art FMs to scale up robot teleoperation by casting it as an economically valuable task that benefits directly from modern large models and agents.

We drive \via~using general-purpose agents, with no robot-specific fine-tuning or additional scaffolding, to demonstrate that, given the right interface and MCP tools, existing agents and models can be strong robot manipulators.
\via~reuses off-the-shelf agents like Claude Code or Codex, including their tool-use loops and context management, so improvements in both the FM and the agent software can transfer to robot manipulation.
We equip the agent with a short system prompt that covers the basics of the interface and provides general guidance on using it effectively, such as the configuration and limitations of the robot, the concept of waypoints and preference for closed-loop verification over open-loop execution (full prompt in \cref{sec:appendix-system-prompt}).

% \subsection{Extensions}
% By leveraging an agentic framework, \via~enables further possible directions for robotics research.

% \textbf{Automatic Tool Improvement.}
% The agent controlling the robot is also a highly capable reasoning and coding agent, so it can suggest modifications to existing tools and implement new tools on its own.
% When we developed the MCP tools, we spun off runs in which an agent read the docstring of a newly defined tool and tested it in the environment.
% We then asked the agent for feedback on both the functionality and the docstring of the tool, and used the feedback to improve them.

% \textbf{Learning via Reflection.}
% Learning from text feedback~\citep{pryzant2023protegi, yuksekgonul2025textgrad} or text reflections~\citep{shinn2023reflexion} has become a popular alternative to reinforcement learning for policy improvement with LLMs, but it has yet to gain broad attention in robotics.
% \via~naturally allows similar ideas to be applied on robot tasks via text or multimodal reflection, without updating model weights.
% We show that it is possible to create task-specific prompts that greatly improve performance, indicating that learning via reflection has a high ceiling for robotics.

\section{Experiments}
\label{sec:experiments}

We evaluate \via~on a suite of six tabletop manipulation tasks with a Franka Panda arm in simulation to demonstrate how \via~scales with different models, compare it with a naive baseline, and examine whether it benefits from detailed text demonstrations.
Then, we run \via~on five real-world manipulation tasks with a Stretch 4 mobile robot to show that the system works well in the real world with different hardware and noisy perception.

% In simulation, we run \via~ with Claude Code (CC) and Codex, each with three model tiers to demonstrate how VIA scales with model capability, and , each with two model tiers and two levels of prompt detail, to assess both the capability of frontier agents, and how their performance scales with model strength.

\subsection{Simulation Experimental Setup}
\label{sec:exp-setup}

\begin{figure}[!t]
\centering
% Source: graphics/task_success.typ; archived successful rollout frames.
\includegraphics[width=\linewidth]{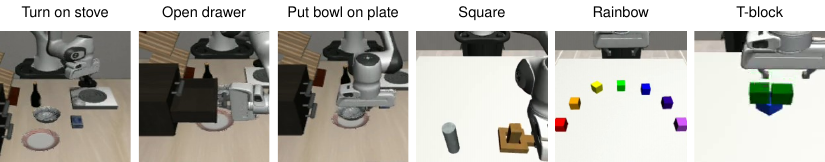}
\caption{\textbf{The six evaluation tasks.}
Each panel shows the final state of a successful \via~rollout, captured by a third-person camera.
Object placements are randomized across seeds.
}
\label{fig:tasks}
\end{figure}
% [adapted: task list from evaluation.md, near-verbatim; robosuite/10-seed framing sentence expanded]
\smallskip\noindent\textbf{Tasks.}
We use the following tasks in our evaluation (\cref{fig:tasks}):

\begin{itemize}[leftmargin=*, itemsep=0pt, parsep=0pt, topsep=2pt, partopsep=0pt]
    \item \emph{Turn on stove}:
    Turn on a stove by rotating its knob, from LIBERO-Goal~\citep{liu2023libero}.
    \item \emph{Open drawer}:
    Open the middle drawer of a three-drawer cabinet, from LIBERO-Goal.
    \item \emph{Put bowl on plate}: Pick up a bowl and place it on the center of the plate, from LIBERO-Goal.
    This demands reasoning on the grasp point and precision, as the bowl must be grasped by pinching its side wall and placed at the exact center of the plate to succeed.
    \item \emph{Square}: Pick up the square nut and fit it onto the square peg, from robosuite~\citep{zhu2020robosuite}.
    \item \emph{Rainbow}: Arrange seven randomly placed colored blocks into a rainbow, a task we created.
    This tests long-horizon planning and execution.
    \item \emph{T-block}: Build a T-shape with three blocks of the same size: one blue block on the bottom and two green blocks resting directly on top of it, side by side; from BuilderBench~\citep{builderbench2025}, re-implemented.
    This requires both solving a physics puzzle and precise manipulation.
\end{itemize}

These tasks test a diverse set of skills: manipulation, physical reasoning, long-term planning, and precise execution.
All tasks run in the robosuite simulator.
We use a simple PI (proportional-integral) controller that linearly interpolates between the current state and the target waypoint. %, but more advanced controllers would be suitable for more complicated domains.
The outcome of each task is a binary success or failure.

% [adapted: evaluation.md, near-verbatim]
\smallskip\noindent\textbf{Evaluation Protocol and Baseline.}
%  model & effort
We use Claude Code (CC) and Codex as our agents.
To evaluate how \via~scales with the size and strength of the underlying model, we run CC with Sonnet 5, Opus 4.8\footnote{Opus 4.8 was the strongest model for a considerable amount of time during
which this project was developed, so we decided to use this model instead of the newer Opus 5 to avoid the cost of re-evaluating.}, and Fable 5, and run Codex with Terra 5.6, Sol 5.6, and Astra 6.
We run all agents at \texttt{xhigh} reasoning effort.
% task prompts
On top of the shared system prompt in \cref{sec:appendix-system-prompt}, we give the model a simple task prompt by default, stating only the task description and success condition but no detailed instructions.
We evaluate each agent-model configuration over 10 seeds per task, with varying initial object placements.
% others
We cap each episode at one hour of wall-clock time, except for Rainbow, which has a two-hour limit, although the agents complete most tasks in far less time when successful.
We disable memory access for the agents so that they cannot accumulate experience across episodes.
We also disable internet search and file access to prevent reading the exact task state specifications from the codebase\footnote{We find that agents may cheat by trying to read the source files of the simulation benchmarks to recover the exact simulation states and object positions if we do not disable general file access.}.

To isolate the benefit of \via, we consider a naive \emph{baseline} that uses the same inputs, agent, and controller setup as \via~but does not have access to the interface.
It receives the same observations (snapshots of the same images and 3D point clouds, along with proprioceptive states) but must issue commands to the controller for immediate execution, without the ability to probe and iterate using any interface components.

\via~unlocks additional opportunities to improve its performance without any gradient-based fine-tuning.
Similar to other agentic systems powered by FMs, we can optimize its prompts or ask the agents to build new tools to improve task performance.
We perform a preliminary case study along this line by giving agents detailed task prompts explaining how we would solve certain tasks with example lists of waypoints.
We refer to this variant as \emph{VIA-detailed}.

\begin{table}[t]
\centering
% Values checked against graphics/data/scaling.json.

\setlength{\tabcolsep}{4pt}
\begin{tabular}{l cccccc}
\toprule
& \multicolumn{3}{c}{Claude Code} & \multicolumn{3}{c}{Codex} \\
\cmidrule(lr){2-4} \cmidrule(lr){5-7}
& Medium & Large & Extra & Medium & Large & Extra \\
Task & Sonnet 5 & Opus 4.8 & Fable 5 & Terra 5.6 & Sol 5.6 & Astra 6 \\
\midrule
Turn on stove & 70\% & 100\% & 100\% & 70\% & 100\% & 100\% \\
Open drawer & 40\% & 70\% & 90\% & 30\% & 80\% & 100\% \\
Put bowl on plate & 50\% & 60\% & 100\% & 70\% & 100\% & 100\% \\
Square & 30\% & 70\% & 80\% & 0\% & 30\% & 100\% \\
Rainbow & 70\% & 80\% & 100\% & 50\% & 100\% & 100\% \\
T-block & 0\% & 10\% & 40\% & 10\% & 30\% & 100\% \\
\midrule
Overall & 43.3\% & 65.0\% & 85.0\% & 38.3\% & 73.3\% & 100.0\% \\
\bottomrule
\end{tabular}%
\caption{
\textbf{Performance of \via~in simulation.}
Success rates over ten seeds per task.
}
\label{tab:scaling-results}
\end{table}

\subsection{Simulation Experimental Results}
\label{sec:exp-results}

% Keep the scaling plot beside the discussion that introduces it.
% Source: graphics/scaling.typ; data: tools/make_scaling_data.py.
\begin{wrapfigure}{R}{0.52\textwidth}
\vspace{-\intextsep}
\centering
\includegraphics[width=\linewidth]{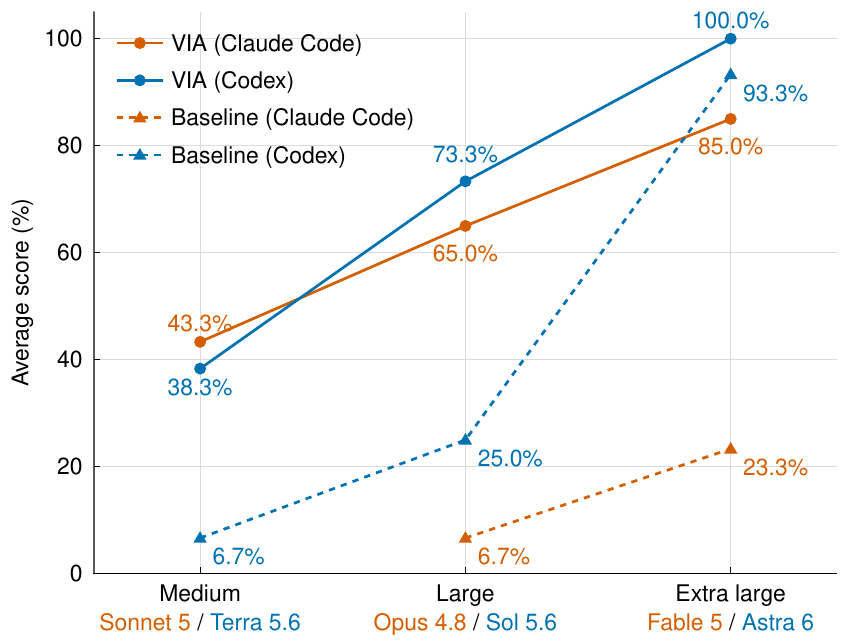}
\caption{\textbf{Performance of \via~and the baseline in simulation.}
\via~scales with the capacity of the underlying models and achieves a perfect score using Astra.
The baseline performs poorly except when using Astra.}
\label{fig:scaling}
\vspace{-2\intextsep} % Tighten the gap before text returns to full width.
\end{wrapfigure}

% [adapted: evaluation.md]
% \cref{tab:main-results} summarizes the success rates of \via~with different configurations across tasks.
\cref{fig:scaling} shows the scaling curve of \via~with different model tiers, measured by average success over all six tasks, as well as a comparison with the baseline.
\cref{tab:scaling-results} shows the per-task success rates for \via, and \cref{tab:scaling-baseline} in \cref{sec:appendix-scaling} shows them for the baseline.
\cref{tab:prompt-results} shows how having detailed instructions in prompts affects performance for different agents.
We report tool call efficiency and estimated API costs in \cref{sec:appendix-efficiency}.

\smallskip\noindent\textbf{\via~solves diverse manipulation tasks zero-shot.}
\via~achieves strong performance zero-shot in our simulation suite, achieving around $40\%$ success even with the smaller models (Sonnet, Terra).
The high success rates on \emph{Turn on stove} and \emph{Open drawer} indicate that \via~is able to handle articulated objects, beyond simple pick-and-place motions.
The \emph{Rainbow} task is specifically designed to test long-horizon planning and execution.
We observe that the agents always plan carefully before moving any block, estimating each block's location and choosing the ordering with the fewest moves.
The effect of such planning is visible in their behavior: they sometimes assemble the rainbow with red on the left and sometimes with red on the right to minimize the number of motions required depending on the initial distribution of the blocks.
\emph{T-block} remains challenging for all except for Codex with Astra.
All agents figure out the right strategy is to rotate the bottom blue block by 45 degrees to increase the contact area, but most of them fail to place the green blocks precisely on each half of the blue block, as this task tolerates only about $8$\,mm of placement error.

\smallskip\noindent\textbf{\via~scales with the capacity of the underlying models.}
\cref{fig:scaling} shows that \via~benefits directly from the scale of the underlying models. %, as it requires no additional fine-tuning and reuses the mature agent harnesses built for these models.
Performance of \via~improves from $43\%$ to $85\%$ by switching to bigger models in Claude Code and from $38\%$ to a perfect $100\%$ for Codex.
% The rate of progress is particularly worth noting, as there have only been less than 5 months between the releases of Opus 4.8, Fable 5 and Astra 6, each substantially pushs the frontier.

\smallskip\noindent\textbf{\via~greatly outperforms the baseline.} %, with Astra 6 being a notable outlier.}
Compared to the naive baseline (dashed lines in \cref{fig:scaling}), which uses the same observations and controller but does not use the interface, \via~performs substantially better across all configurations except for Astra 6, where the interface improves performance only slightly, from $93\%$ to $100\%$.
The surprisingly strong performance of the baseline with Astra 6 is also consistent with many recent demonstrations showing that Astra 6 can perform many robot manipulation tasks zero-shot.
Our evaluation quantifies this step change by comparing performance across models, highlighting how poorly prior state-of-the-art models perform without \via.

%  \via~is crucial for all the non-Astra models to get a good performance in these tasks, with performance dropping significantly .

% \smallskip\noindent\textbf{\via~scales with model size and capabilities.}
% Within the CC family, upgrading from Opus 4.8 to Fable 5 improves the success rate on every task that has remaining headroom, raising the overall success rate from $70\%$ to $88\%$.
% On the LIBERO-Goal tasks, CC-Fable passes 29/30 seeds with only the minimal prompt; in fact, its single failure would have been avoided had it pulled the drawer a few more centimeters out.
% With the minimal prompt alone, it essentially matches the perfect score that CC-Opus achieves only with the detailed instructions.
% The GPT family moves in the same direction on the precision-demanding \emph{Put bowl on plate}, where upgrading from GPT-5.5 to GPT-5.6-Sol doubles the success rate from $40\%$ to $80\%$.
% Its results elsewhere are mixed: \emph{Rainbow} is roughly unchanged ($60\%$ to $50\%$), and \emph{T-block} drops from $40\%$ to $20\%$.
% This mirrors a broader trend in foundation model research: as models improve, less prompt engineering is needed to reach the same performance.
% As general computer use becomes a centerpiece of recent model releases~\citep{anthropic2026fable, openai2026gpt56}, \via~is well-positioned to harvest these general gains for robot control.

\begin{table}[t]
\centering
\setlength{\tabcolsep}{4pt}
\begin{tabular}{l cc cc}
\toprule
& \multicolumn{2}{c}{Claude Code Opus 4.8} & \multicolumn{2}{c}{Codex Sol 5.6} \\
\cmidrule(lr){2-3} \cmidrule(lr){4-5}
Task & Default & Detailed & Default & Detailed \\
\midrule
Turn on stove     & \textbf{100\%} & \textbf{100\%} & \textbf{100\%} & 80\% \\
Open drawer       & 70\%           & \textbf{100\%} & \textbf{80\%} & 70\% \\
Put bowl on plate & 60\%           & \textbf{100\%} & \textbf{100\%} & \textbf{100\%} \\
Square            & \textbf{70\%}  & 60\%           & 30\%  & \textbf{80\%} \\
T-block           & 10\%           & \textbf{20\%}  & 30\%  & \textbf{40\%} \\
\midrule
Avg.              & 62\%           & \textbf{76\%}  & 68\%  & \textbf{74\%} \\
\bottomrule
\end{tabular}
\caption{\textbf{Default versus detailed prompts.}
Success rates of \via~under default and detailed prompt variants, each evaluated over 10 episodes per task.
Bold marks the higher success rate, including ties.
Detailed prompts bring moderate improvements by supplying the agents with lists of example waypoints written by humans.
}
\label{tab:prompt-results}
\vspace{-8pt} % Tighten the gap before the real-world figure.
\end{table}

\smallskip\noindent\textbf{Detailed text instructions bring modest improvement.}
In \cref{tab:prompt-results}, we compare default and detailed prompts on five simulation tasks, excluding Rainbow, using the \emph{large} configurations from \cref{fig:scaling}.
We manually write detailed instructions for each of these tasks explaining how we would complete the task using this interface and feed the instructions to the agents as additional prompts.
Overall, the instructions help modestly, improving the success rate by $14\%$ for CC Opus and $6\%$ for Codex Sol.
The remaining failure modes vary from task to task.
For example, T-block failures are primarily due to imprecise perception and execution; Codex Sol's failures in \emph{drawer} are caused by opening the wrong drawer, which is not specifically emphasized in the detailed instructions.
A more promising direction may be to have the agents review their own failures and optimize prompts accordingly instead of having a human write them.

% We equip CC-Opus and Codex-5.5, the weaker model from each family, with the detailed prompts that provide example waypoint lists for the three LIBERO-Goal tasks.
% CC-Opus improves significantly on these three tasks, from $77\%$ to a perfect $100\%$, while Codex-5.5 does not benefit from them.
% The improvement of CC-Opus is encouraging, as it shows that \via~offers a low-cost way to create ``demonstrations'' for robots from plain text or screenshots, rather than teleoperation.
% It also opens up the possibility of applying automatic prompt optimization so that agents can improve their own task prompts from past experience, i.e., learning via reflection.

\subsection{Real-World Evaluation}
\label{sec:real-world-eval}

\begin{figure}[t]
\centering
% Source: graphics/real_task_annotations.py; archived photographs and editable motion annotations.
\includegraphics[width=\linewidth]{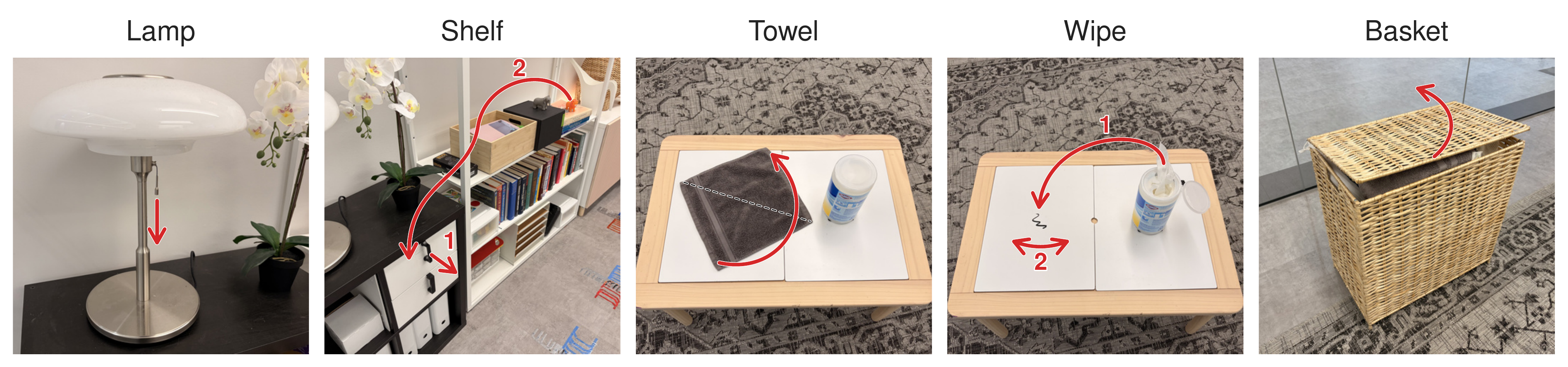}
\caption{\textbf{The five real-world evaluation tasks.}
Red arrows indicate the intended motions.
Initial locations of the objects and the mobile Stretch 4 robot are both randomized for each episode.
}
\label{fig:real-world-tasks}
\vspace{4pt}
% Table 3 lives in the same float so it always appears next to the figure.
\setlength{\tabcolsep}{8pt}
\begin{tabular}{l ccccc}
\toprule
Method & Lamp & Shelf & Towel & Wipe & Basket \\
\midrule
\via & 4/5 & 5/5 & 5/5 & 5/5 & 5/5 \\
\bottomrule
\end{tabular}
\captionof{table}{\textbf{Real-world evaluation.}
Success rates of \via~(w/ Codex Astra 6) on real-world tasks.}
\label{tab:real-world-results}
\vspace{-\intextsep} % Tighten the gap below the combined figure and table.
\end{figure}

For the real robot experiments, instead of staying in the stationary tabletop domain, we use a mobile manipulator, Stretch 4 (illustrations and details in \cref{sec:appendix-real-robot}), that operates at room scale to demonstrate that the system works with coarse and noisy depth perception.
The mobile robot also allows for a much wider range of random initial conditions, as well as multi-step tasks that require the agent to remember objects that go out of sight as it moves the robot.
Unlike in simulation, where we have perfect depth information to construct dense, informative point clouds, the Stretch 4 robot has a head-mounted lidar sensor that provides depth information only for sparse points at room scale.
Therefore, we use a collection of purely 2D-based UI components for the interface (illustrations in \cref{fig:interface-interactions} (c) and (d)) but allow the agents to probe depth with \texttt{hover}.
We run Fast-FoundationStereo~\citep{wen2026fastfoundationstereo} for the stereo camera on the wrist to provide better depth estimation in the wrist view.

\smallskip\noindent\textbf{Tasks.}
We use the following tasks in our real-world evaluation (\cref{fig:real-world-tasks}).
Apart from the task-specific randomization described below, the initial location of the robot is also randomized.
\begin{itemize}[leftmargin=*, itemsep=0pt, parsep=0pt, topsep=2pt, partopsep=0pt]
    \item \emph{Lamp}: Turn on the lamp by pulling the small chain downward. The lamp may not appear in the initial view of the robot. The chain is small and requires precise motion.
    \item \emph{Shelf}: Open the top drawer and put the toy lion from the shelf into the drawer. The toy lion's location and pose are randomized over the shelf. The toy and drawer are far apart so that they do not appear in the robot's view at the same time.
    \item \emph{Towel}: Fold a towel diagonally by bringing one corner to the opposite corner. The location and pose of both the towel and the table are randomized.
    This tests deformable object manipulation.
    \item \emph{Wipe}: Pull a wipe out of its container and use it to clean the black mark on the table. The location and shape of the black mark, as well as the location and orientation of the table, are randomized.
    \item \emph{Basket}: Open the basket until the lid flips to the back of the basket. The location and orientation of the basket are randomized. This manipulation of an articulated object requires complex motion.
\end{itemize}

\smallskip\noindent\textbf{Results.}
\cref{tab:real-world-results} summarizes the results.
We evaluate \via~with Codex Astra 6 as the agent for five episodes in each task, and it achieves a near-perfect $24/25$ ($96\%$) success rate.
% These results show that \via~reliably extends to realistic real-world scenarios with commercial hardware.
These results show that \via~extends to real-world manipulation with commercial hardware.
The \emph{Towel} and \emph{Wipe} tasks extend the evaluation suite to deformable objects.
Apart from the quantitative results, these evaluations also exhibit several interesting behaviors.
In \emph{Lamp}, the task description mentions that the lamp is on a cabinet, and the agent is able to search for the lamp quickly once the cabinet appears in its view.
The one failure in this task is when the agent overextends the gripper to hold the stem of the lamp instead of the chain, which could be a perception problem as the chain and the stem are the same silver color.
% In \emph{Shelf}, the limited viewport of the head camera means that the agent has to memorize the locations of the lion and drawer to move efficiently between the two, and \via~is able to do it without any additional engineering from us thanks to the built-in context management of the underlying agent.
In \emph{Shelf}, the head camera's limited field of view requires the agent to remember the locations of the lion and drawer to move efficiently between them.
\via~does this using the underlying agent's built-in context management, without any additional engineering.
In \emph{Towel}, the agent is able to fold the towel precisely so that the edges align well.
In \emph{Wipe}, the agent's view is blocked when wiping, and it automatically moves the gripper away after a few wipes to check the cleanliness of the table.
In one episode, it drops the wipe on the table and pushes it with a closed gripper instead of holding it, after the latter strategy fails to wipe cleanly.
% Finally, in \emph{Basket}, the agents automatically retry within episodes when the lid falls off during opening.
Finally, in \emph{Basket}, the agent automatically retries within episodes when the lid slips out of the gripper during opening.

\section{Conclusion}
\label{sec:conclusion}

We present \via, a framework that recasts robot control as a visual agentic task: an off-the-shelf FM-powered agent operates an interface built from RGB-D streams, posing target grippers and executing waypoints through human-legible MCP tools.
Using off-the-shelf frontier agents, \via~solves six simulated manipulation tasks and succeeds in $24/25$ trials across five real-world mobile manipulation tasks, without robot-specific fine-tuning or privileged state.
Performance improves with model strength: each new generation of general capability transfers to the robot, with no fine-tuning required.
One direction for future work is to include a more sophisticated controller, which could allow the system to avoid collisions or handle dynamic tasks.
Another direction is to study how to extend similar ideas to dexterous manipulation beyond parallel grippers.
Even in its current form, however, \via~has practical uses: assembly tasks where reliability matters more than speed, reset mechanisms for real-world reinforcement learning, and data collection in which a slow interface agent generates demonstrations for training fast policies.

\smallskip\noindent\textbf{Limitations.}
\via~achieves high success rates with the frontier agents, but inference remains slow and expensive.
We expect similar performance from smaller and cheaper models as FMs continue to improve, with open-weight or less congested models further reducing cost and latency.
Slow inference also restricts \via~to quasi-static tasks: it is not suitable for dynamic tasks like catching a ball.
However, a wide range of quasi-static tasks carry significant value and existing methods are not yet reliable on them, making \via~a compelling option in this regime.

% ./
% \smallskip\noindent\textbf{Future Work.}
% The most immediate next step is to extend \via~to real robots.

% funding sources go here

% \newpage
\section*{Acknowledgments}
We would like to thank Michelle Pan, Suvir Mirchandani and Omar Shaikh for insightful discussions, and thank Michelle Pan and Zen Yaskawa for help with the Stretch robot.

\section*{AI use statement}
In this work, we used generative AI tools for assistance with code implementation, running experiments, creating figures, and polishing the manuscript.
The core idea, the method, and the experiments are all conceived and designed by the authors.
We have reviewed all AI-assisted work.
We take responsibility for the final content of this work, including text, claims or artifacts produced with the aid of generative AI.

% \subsection*{Ethics statement}
% This work tackles applications of foundation models and agents in robotic control.
% We are not aware of any ethical issues regarding this work.

\section*{Reproducibility statement}
We have provided details to reproduce our experiments, including documentation of our MCP tools and prompts, in \cref{sec:appendix-tools} and \cref{sec:appendix-prompts}.
We also plan to fully open source the code and video recordings of the evaluation episodes.

\bibliography{references}

@techreport{radford2019language,
  title       = {{Language Models are Unsupervised Multitask Learners}},
  author      = {Radford, Alec and Wu, Jeffrey and Child, Rewon and Luan, David and Amodei, Dario and Sutskever, Ilya},
  institution = {OpenAI},
  year        = {2019},
}

@article{bommasani2021foundation,
  title   = {{On the Opportunities and Risks of Foundation Models}},
  author  = {Bommasani, Rishi and others},
  journal = {arXiv preprint arXiv:2108.07258},
  year    = {2021},
}

@article{openai2023gpt4,
  title   = {{{GPT-4} Technical Report}},
  author  = {{OpenAI}},
  journal = {arXiv preprint arXiv:2303.08774},
  year    = {2023},
}

@inproceedings{yao2023react,
  title     = {{{ReAct}: Synergizing Reasoning and Acting in Language Models}},
  author    = {Yao, Shunyu and Zhao, Jeffrey and Yu, Dian and Du, Nan and Shafran, Izhak and Narasimhan, Karthik and Cao, Yuan},
  booktitle = {International Conference on Learning Representations (ICLR)},
  year      = {2023},
}

@inproceedings{brohan2023rt2,
  title     = {{{RT-2}: Vision-Language-Action Models Transfer Web Knowledge to Robotic Control}},
  author    = {Zitkovich, Brianna and others},
  booktitle = {Conference on Robot Learning (CoRL)},
  year      = {2023},
}

@inproceedings{kim2024openvla,
  title     = {{{OpenVLA}: An Open-Source Vision-Language-Action Model}},
  author    = {Kim, Moo Jin and Pertsch, Karl and Karamcheti, Siddharth and Xiao, Ted and Balakrishna, Ashwin and Nair, Suraj and Rafailov, Rafael and Foster, Ethan P. and Sanketi, Pannag R. and Vuong, Quan and Kollar, Thomas and Burchfiel, Benjamin and Tedrake, Russ and Sadigh, Dorsa and Levine, Sergey and Liang, Percy and Finn, Chelsea},
  booktitle = {Conference on Robot Learning (CoRL)},
  year      = {2024},
}

@inproceedings{black2024pi0,
  title     = {{$\pi_0$: A {Vision-Language-Action} Flow Model for General Robot Control}},
  author    = {Black, Kevin and Brown, Noah and Driess, Danny and Esmail, Adnan and Equi, Michael Robert and Finn, Chelsea and Fusai, Niccolo and Groom, Lachy and Hausman, Karol and Ichter, Brian and Jakubczak, Szymon and Jones, Tim and Ke, Liyiming and Levine, Sergey and Li-Bell, Adrian and Mothukuri, Mohith and Nair, Suraj and Pertsch, Karl and Shi, Lucy Xiaoyang and Smith, Laura and Tanner, James and Vuong, Quan and Walling, Anna and Wang, Haohuan and Zhilinsky, Ury},
  booktitle = {Robotics: Science and Systems (RSS)},
  year      = {2025},
}

@inproceedings{fu2026capx,
  title   = {{{CaP-X}: A Framework for Benchmarking and Improving Coding Agents for Robot Manipulation}},
  author  = {Fu, Letian and Yu, Justin and El-Refai, Karim and Kou, Ethan and Xue, Haoru and Huang, Huang and Xiao, Wenli and Li, Fei-Fei and Shi, Guanya and Wu, Jiajun and Sastry, S. Shankar and Zhu, Yuke and Goldberg, Ken and Fan, Linxi},
  booktitle = {International Conference on Machine Learning (ICML)},
  year      = {2026},
}

@inproceedings{hancock2025vlm2vla,
  title     = {{Actions as Language: Fine-Tuning {VLMs} into {VLAs} Without Catastrophic Forgetting}},
  author    = {Hancock, Asher James and Wu, Xindi and Zha, Lihan and Russakovsky, Olga and Majumdar, Anirudha},
  booktitle = {International Conference on Learning Representations (ICLR)},
  year      = {2026},
}

@inproceedings{liang2023code,
  title     = {{Code as Policies: Language Model Programs for Embodied Control}},
  author    = {Liang, Jacky and Huang, Wenlong and Xia, Fei and Xu, Peng and Hausman, Karol and Ichter, Brian and Florence, Pete and Zeng, Andy},
  booktitle = {IEEE International Conference on Robotics and Automation (ICRA)},
  year      = {2023},
}

@inproceedings{singh2023progprompt,
  title     = {{{ProgPrompt}: Generating Situated Robot Task Plans using Large Language Models}},
  author    = {Singh, Ishika and Blukis, Valts and Mousavian, Arsalan and Goyal, Ankit and Xu, Danfei and Tremblay, Jonathan and Fox, Dieter and Thomason, Jesse and Garg, Animesh},
  booktitle = {IEEE International Conference on Robotics and Automation (ICRA)},
  year      = {2023},
}

@inproceedings{ahn2022saycan,
  title     = {{Do As I Can, Not As I Say: Grounding Language in Robotic Affordances}},
  author    = {ichter, brian and Brohan, Anthony and Chebotar, Yevgen and Finn, Chelsea and Hausman, Karol and Herzog, Alexander and Ho, Daniel and Ibarz, Julian and Irpan, Alex and Jang, Eric and Julian, Ryan and Kalashnikov, Dmitry and Levine, Sergey and Lu, Yao and Parada, Carolina and Rao, Kanishka and Sermanet, Pierre and Toshev, Alexander T and Vanhoucke, Vincent and Xia, Fei and Xiao, Ted and Xu, Peng and Yan, Mengyuan and Brown, Noah and Ahn, Michael and Cortes, Omar and Sievers, Nicolas and Tan, Clayton and Xu, Sichun and Reyes, Diego and Rettinghouse, Jarek and Quiambao, Jornell and Pastor, Peter and Luu, Linda and Lee, Kuang-Huei and Kuang, Yuheng and Jesmonth, Sally and Joshi, Nikhil J. and Jeffrey, Kyle and Ruano, Rosario Jauregui and Hsu, Jasmine and Gopalakrishnan, Keerthana and David, Byron and Zeng, Andy and Fu, Chuyuan Kelly},
  booktitle = {Conference on Robot Learning (CoRL)},
  pages     = {287--318},
  year      = {2023},
  url       = {https://proceedings.mlr.press/v205/ichter23a.html},
}

@article{lin2023text2motion,
  title   = {{Text2Motion: From Natural Language Instructions to Feasible Plans}},
  author  = {Lin, Kevin and Agia, Christopher and Migimatsu, Toki and Pavone, Marco and Bohg, Jeannette},
  journal = {Autonomous Robots},
  year    = {2023},
  url     = {https://arxiv.org/abs/2303.12153},
}

@inproceedings{debakker2025scaffolding,
  title     = {{Scaffolding Dexterous Manipulation with Vision-Language Models}},
  author    = {de Bakker, Vincent and Hejna, Joey and Lum, Tyler and Celik, Onur and Taranovic, Aleksandar and Blessing, Denis and Neumann, Gerhard and Bohg, Jeannette and Sadigh, Dorsa},
  booktitle = {Advances in Neural Information Processing Systems (NeurIPS)},
  pages     = {93317--93352},
  year      = {2025},
  doi       = {10.52202/085713-3121},
  url       = {https://proceedings.neurips.cc/paper_files/paper/2025/hash/862644b156e51c35dea5a7446d640b14-Abstract-Conference.html},
}

@inproceedings{shi2025hirobot,
  title     = {{{Hi Robot}: Open-Ended Instruction Following with Hierarchical Vision-Language-Action Models}},
  author    = {Shi, Lucy Xiaoyang and Ichter, Brian and Equi, Michael and Ke, Liyiming and Pertsch, Karl and Vuong, Quan and Tanner, James and Walling, Anna and Wang, Haohuan and Fusai, Niccolo and Li-Bell, Adrian and Driess, Danny and Groom, Lachy and Levine, Sergey and Finn, Chelsea},
  booktitle = {International Conference on Machine Learning (ICML)},
  year      = {2025},
  url       = {https://arxiv.org/abs/2502.19417},
}

@inproceedings{li2025hamster,
  title     = {{{HAMSTER}: Hierarchical Action Models for Open-World Robot Manipulation}},
  author    = {Li, Yi and Deng, Yuquan and Zhang, Jesse and Jang, Joel and Memmel, Marius and Garrett, Caelan and Ramos, Fabio and Fox, Dieter and Li, Anqi and Gupta, Abhishek and Goyal, Ankit},
  booktitle = {International Conference on Learning Representations (ICLR)},
  pages     = {24040--24068},
  year      = {2025},
  url       = {https://proceedings.iclr.cc/paper_files/paper/2025/hash/3bfee3bc6639c36e6e7b058db909f760-Abstract-Conference.html},
}

@inproceedings{huang2025thinkact,
  title     = {{{ThinkAct}: Vision-Language-Action Reasoning via Reinforced Visual Latent Planning}},
  author    = {Huang, Chi-Pin and Wu, Yueh-Hua and Chen, Min-Hung and Wang, Yu-Chiang Frank and Yang, Fu-En},
  booktitle = {Advances in Neural Information Processing Systems (NeurIPS)},
  year      = {2025},
  url       = {https://arxiv.org/abs/2507.16815},
}

@article{chen2026volo,
  title   = {{{VoLo}: A Physical Orchestrator for Open-Vocabulary Long-Horizon Manipulation}},
  author  = {Chen, Siyi and Hadfield, Hugo and Zook, Alex and Uy, Mikaela Angelina and Song, Chan Hee and Coumans, Erwin and Yang, Xuning and Ladhak, Faisal and Qu, Qing and Birchfield, Stan and Tremblay, Jonathan and Blukis, Valts},
  journal = {arXiv preprint arXiv:2606.07723},
  year    = {2026},
  url     = {https://arxiv.org/abs/2606.07723},
}

@article{huang2023instruct2act,
  title   = {{{Instruct2Act}: Mapping Multi-modality Instructions to Robotic Actions with Large Language Model}},
  author  = {Huang, Siyuan and Jiang, Zhengkai and Dong, Hao and Qiao, Yu and Gao, Peng and Li, Hongsheng},
  journal = {arXiv preprint arXiv:2305.11176},
  year    = {2023},
  url     = {https://arxiv.org/abs/2305.11176},
}

@inproceedings{huang2023voxposer,
  title     = {{{VoxPoser}: Composable {3D} Value Maps for Robotic Manipulation with Language Models}},
  author    = {Huang, Wenlong and Wang, Chen and Zhang, Ruohan and Li, Yunzhu and Wu, Jiajun and Fei-Fei, Li},
  booktitle = {Conference on Robot Learning (CoRL)},
  year      = {2023},
  url       = {https://arxiv.org/abs/2307.05973},
}

@inproceedings{nasiriany2024pivot,
  title     = {{{PIVOT}: Iterative Visual Prompting Elicits Actionable Knowledge for {VLMs}}},
  author    = {Nasiriany, Soroush and Xia, Fei and Yu, Wenhao and Xiao, Ted and Liang, Jacky and Dasgupta, Ishita and Xie, Annie and Driess, Danny and Wahid, Ayzaan and Xu, Zhuo and Vuong, Quan and Zhang, Tingnan and Lee, Tsang-Wei Edward and Lee, Kuang-Huei and Xu, Peng and Kirmani, Sean and Zhu, Yuke and Zeng, Andy and Hausman, Karol and Heess, Nicolas and Finn, Chelsea and Levine, Sergey and Ichter, Brian},
  booktitle = {International Conference on Machine Learning (ICML)},
  pages     = {37321--37341},
  year      = {2024},
  url       = {https://proceedings.mlr.press/v235/nasiriany24a.html},
}

@inproceedings{liu2024moka,
  title     = {{{MOKA}: Open-World Robotic Manipulation through Mark-Based Visual Prompting}},
  author    = {Fang, Kuan and Liu, Fangchen and Abbeel, Pieter and Levine, Sergey},
  booktitle = {Robotics: Science and Systems (RSS)},
  year      = {2024},
  doi       = {10.15607/RSS.2024.XX.062},
  url       = {https://www.roboticsproceedings.org/rss20/p062.html},
}

@inproceedings{shi2017worldofbits,
  title     = {{{World of Bits}: An Open-Domain Platform for Web-Based Agents}},
  author    = {Shi, Tianlin and Karpathy, Andrej and Fan, Linxi and Hernandez, Jonathan and Liang, Percy},
  booktitle = {International Conference on Machine Learning (ICML)},
  year      = {2017},
}

@misc{anthropic2024computeruse,
  title        = {{Introducing computer use, a new {Claude 3.5 Sonnet}, and {Claude 3.5 Haiku}}},
  author       = {{Anthropic}},
  year         = {2024},
  howpublished = {\url{https://www.anthropic.com/news/3-5-models-and-computer-use}}
}

@misc{openai2025operator,
  title        = {{Introducing {Operator}}},
  author       = {{OpenAI}},
  year         = {2025},
  howpublished = {\url{https://openai.com/index/introducing-operator/}}
}

@misc{openai2025codex,
  title        = {{Introducing {Codex}}},
  author       = {{OpenAI}},
  year         = {2025},
  howpublished = {\url{https://openai.com/index/introducing-codex/}}
}

@misc{claudecode2025,
  title        = {{{Claude Code}}},
  author       = {{Anthropic}},
  year         = {2025},
  howpublished = {\url{https://claude.com/product/claude-code}}
}

@misc{anthropic2024mcp,
  title        = {{Introducing the {Model Context Protocol}}},
  author       = {{Anthropic}},
  year         = {2024},
  howpublished = {\url{https://www.anthropic.com/news/model-context-protocol}}
}

@inproceedings{xie2024osworld,
  title     = {{{OSWorld}: Benchmarking Multimodal Agents for Open-Ended Tasks in Real Computer Environments}},
  author    = {Xie, Tianbao and Zhang, Danyang and Chen, Jixuan and Li, Xiaochuan and Zhao, Siheng and Cao, Ruisheng and Hua, Toh Jing and Cheng, Zhoujun and Shin, Dongchan and Lei, Fangyu and Liu, Yitao and Xu, Yiheng and Zhou, Shuyan and Savarese, Silvio and Xiong, Caiming and Zhong, Victor and Yu, Tao},
  booktitle = {Advances in Neural Information Processing Systems (NeurIPS), Datasets and Benchmarks Track},
  year      = {2024},
}

@inproceedings{zhou2024webarena,
  title     = {{{WebArena}: A Realistic Web Environment for Building Autonomous Agents}},
  author    = {Zhou, Shuyan and Xu, Frank F. and Zhu, Hao and Zhou, Xuhui and Lo, Robert and Sridhar, Abishek and Cheng, Xianyi and Ou, Tianyue and Bisk, Yonatan and Fried, Daniel and Alon, Uri and Neubig, Graham},
  booktitle = {International Conference on Learning Representations (ICLR)},
  year      = {2024},
}

@inproceedings{jimenez2024swebench,
  title     = {{{SWE-bench}: Can Language Models Resolve Real-World {GitHub} Issues?}},
  author    = {Jimenez, Carlos E. and Yang, John and Wettig, Alexander and Yao, Shunyu and Pei, Kexin and Press, Ofir and Narasimhan, Karthik},
  booktitle = {International Conference on Learning Representations (ICLR)},
  year      = {2024},
}

@inproceedings{yang2024sweagent,
  title     = {{{SWE-agent}: Agent-Computer Interfaces Enable Automated Software Engineering}},
  author    = {Yang, John and Jimenez, Carlos E. and Wettig, Alexander and Lieret, Kilian and Yao, Shunyu and Narasimhan, Karthik and Press, Ofir},
  booktitle = {Advances in Neural Information Processing Systems (NeurIPS)},
  year      = {2024},
}

@misc{anthropic2025claude37,
  title        = {{{Claude 3.7 Sonnet} and {Claude Code}}},
  author       = {{Anthropic}},
  year         = {2025},
  howpublished = {\url{https://www.anthropic.com/news/claude-3-7-sonnet}},
}

@inproceedings{shridhar2023peract,
  title     = {{Perceiver-Actor: A Multi-Task Transformer for Robotic Manipulation}},
  author    = {Shridhar, Mohit and Manuelli, Lucas and Fox, Dieter},
  booktitle = {Conference on Robot Learning (CoRL)},
  pages     = {785--799},
  year      = {2023},
  url       = {https://proceedings.mlr.press/v205/shridhar23a.html},
}

@inproceedings{goyal2023rvt,
  title     = {{{RVT}: Robotic View Transformer for {3D} Object Manipulation}},
  author    = {Goyal, Ankit and Xu, Jie and Guo, Yijie and Blukis, Valts and Chao, Yu-Wei and Fox, Dieter},
  booktitle = {Conference on Robot Learning (CoRL)},
  pages     = {694--710},
  year      = {2023},
  url       = {https://proceedings.mlr.press/v229/goyal23a.html},
}

@inproceedings{belkhale2023hydra,
  title     = {{{HYDRA}: Hybrid Robot Actions for Imitation Learning}},
  author    = {Belkhale, Suneel and Cui, Yuchen and Sadigh, Dorsa},
  booktitle = {Conference on Robot Learning (CoRL)},
  pages     = {2113--2133},
  year      = {2023},
  url       = {https://proceedings.mlr.press/v229/belkhale23a.html},
}

@inproceedings{sundaresan2025sphinx,
  title     = {{What's the Move? Hybrid Imitation Learning via Salient Points}},
  author    = {Sundaresan, Priya and Hu, Hengyuan and Vuong, Quan and Bohg, Jeannette and Sadigh, Dorsa},
  booktitle = {International Conference on Learning Representations (ICLR)},
  year      = {2025},
}

@article{zhu2020robosuite,
  title   = {{{robosuite}: A Modular Simulation Framework and Benchmark for Robot Learning}},
  author  = {Zhu, Yuke and Wong, Josiah and Mandlekar, Ajay and Mart{\'i}n-Mart{\'i}n, Roberto and Joshi, Abhishek and Lin, Kevin and Maddukuri, Abhiram and Nasiriany, Soroush and Zhu, Yifeng},
  journal = {arXiv preprint arXiv:2009.12293},
  year    = {2020},
}

@inproceedings{liu2023libero,
  title     = {{{LIBERO}: Benchmarking Knowledge Transfer for Lifelong Robot Learning}},
  author    = {Liu, Bo and Zhu, Yifeng and Gao, Chongkai and Feng, Yihao and Liu, Qiang and Zhu, Yuke and Stone, Peter},
  booktitle = {Advances in Neural Information Processing Systems (NeurIPS), Datasets and Benchmarks Track},
  year      = {2023},
}

@article{builderbench2025,
  title   = {{{BuilderBench}: The Building Blocks of Intelligent Agents}},
  author  = {Ghugare, Raj and Creus Castanyer, Roger and Ji, Catherine and Wantlin, Kathryn and Schofield, Jin and Narasimhan, Karthik and Eysenbach, Benjamin},
  journal = {arXiv preprint arXiv:2510.06288},
  year    = {2025},
}

@inproceedings{wen2026fastfoundationstereo,
  title     = {{{Fast-FoundationStereo}: Real-Time Zero-Shot Stereo Matching}},
  author    = {Wen, Bowen and Dewan, Shaurya and Birchfield, Stan},
  booktitle = {Proceedings of the IEEE/CVF Conference on Computer Vision and Pattern Recognition (CVPR)},
  pages     = {7513--7524},
  year      = {2026},
  url       = {https://nvlabs.github.io/Fast-FoundationStereo/},
}

@misc{hellorobot2026stretch4,
  title        = {{{Stretch 4}: Mobile Manipulator Data Sheet}},
  author       = {{Hello Robot}},
  year         = {2026},
  howpublished = {\url{https://hello-robot.com/wp-content/uploads/2026/05/HelloRobot-DataSheet-Stretch-4-Rev5_AsLaunched.pdf}},
}
\bibliographystyle{iclr2026_conference}

\clearpage
\appendix
% Start each top-level appendix on a fresh page; let subsections flow naturally.
% \input{tex/A3_setups} % comparison table omitted for now
\section{Documentation of MCP Tools}
\label[appendix]{sec:appendix-tools}
\label{sec:method-tools}

This appendix includes the detailed signatures and docstrings of the \via~MCP tools for the simulation and real-world experiments.
These docstrings are accessible to both Claude Code and Codex when they connect to the MCP server.
Square brackets in a signature mark optional arguments.

\subsection{Simulation Tools}
\label{sec:appendix-sim-tools}

% Keep the table below this heading; allow top placement on a later page.
\suppressfloats[t]
\begin{table}[ht]
\centering
\renewcommand{\arraystretch}{1.1}
\begin{tabular}{@{}l >{\raggedright\arraybackslash}p{10.9cm}@{}}
\toprule
\textbf{Category} & \textbf{Tools} \\
\midrule
Observation & \texttt{screenshot}, \texttt{hover}, \texttt{gripper\_get\_pose}, \newline \texttt{gripper\_show\_rotation\_gizmo}, \texttt{camera\_get\_pose} \\
\addlinespace[3pt]
Action & \texttt{gripper\_teleport\_via\_click}, \texttt{gripper\_drag}, \texttt{gripper\_translate}, \texttt{gripper\_advance\_or\_retreat}, \texttt{gripper\_rotate}, \texttt{gripper\_toggle}, \texttt{gripper\_reset} \\
\addlinespace[3pt]
Camera & \texttt{camera\_orbit\_via\_key}, \texttt{camera\_pan\_via\_key}, \texttt{camera\_zoom}, \texttt{camera\_reset} \\
\addlinespace[3pt]
Execution & \texttt{execute\_waypoint}, \texttt{end\_episode} \\
\bottomrule
\end{tabular}
\caption{\textbf{MCP tools for simulation.}
The agent perceives the interface with \texttt{screenshot} and poses the blue target gripper with intuitive tools such as \texttt{gripper\_teleport\_via\_click}, \texttt{gripper\_translate}, and \texttt{gripper\_rotate}.
It can also move the camera via the camera tools to disambiguate the scene.
\texttt{execute\_waypoint} moves the real gripper to the target pose through a controller.
Most tools return an updated screenshot reflecting their effect on the UI or the real environment.
}
\label{tab:tools}
\end{table}

The UI described in \cref{sec:method-interface} already enables a human to control the robot.
To make it operable by agents, we wrap its core interactions into a small set of MCP tools, summarized in \cref{tab:tools} and documented in full below.
The tools fall into four categories.
\emph{Observation tools} gather information about the environment.
\texttt{screenshot} returns a screenshot of the entire UI.
\texttt{gripper\_get\_pose} and \texttt{camera\_get\_pose} return the pose of the virtual target gripper that the agent is commanding and the current viewport parameters, respectively.
We also provide two helper tools.
\texttt{hover} takes a pair of normalized screenshot coordinates \texttt{(u, v)} and returns a screenshot with a virtual crosshair overlaid at that location, along with the 3D coordinates of the point it hits in the point cloud, if any.
The agent can use it to verify that a planned click will land at the intended location, or to read off the 3D coordinates of points of interest.
\texttt{gripper\_show\_rotation\_gizmo} overlays a rotation gizmo on the target gripper to help the agent reason about rotation directions.
\emph{Action tools} manipulate the target gripper.
They have no effect on the real robot until \texttt{execute\_waypoint} is called.
Each action tool returns a screenshot after its change to the target gripper is rendered on the UI, so the agent can check the outcome and keep adjusting until the waypoint is ready for execution.
\emph{Camera tools} change the viewport so that the agent can observe the scene from angles beyond the default view.
\emph{Execution tools} affect the underlying environment, either moving the robot via \texttt{execute\_waypoint}, or terminating the episode via \texttt{end\_episode}.

We design the tools around two principles: \emph{minimalism} and \emph{agent ergonomics}.
For minimalism, each tool wraps the corresponding human operation with little or no extra abstraction, so agents can operate the system with the same control and freedom as a human.
For agent ergonomics, we account for the fact that agents observe the UI through discrete screenshots rather than continuous animation.
Operations such as rotating the gripper by dragging a gizmo axis or orbiting the camera with a continuous drag are therefore hard for an agent to comprehend, so we provide more direct alternatives, such as \texttt{gripper\_rotate} and \texttt{camera\_orbit\_via\_key}, which take exact angles as input.

% Shaded card per tool: darker title strip with the signature, light shaded body.
\newtcolorbox{tooldoc}[1]{
  enhanced, breakable,
  frame hidden, boxrule=0pt, arc=3pt,
  colback=gray!8,
  colbacktitle=gray!25, coltitle=black, fonttitle=\ttfamily,
  title={#1},
  left=6pt, right=6pt, top=5pt, bottom=5pt,
  toptitle=3pt, bottomtitle=3pt,
  before skip=10pt, after skip=8pt,
}

\begingroup
\setlist[itemize]{leftmargin=1.5em, itemsep=1pt, topsep=2pt, parsep=0pt}

All gripper tools except \texttt{gripper\_toggle} also share a common preamble describing the coordinate frame of the system, reproduced below.

\begin{tooldoc}{Shared coordinate-frame preamble}
Frame (shared by all spatial tools):
\begin{itemize}
\item \emph{robot}: metres; the canonical frame every pose, delta, and camera target reports in and that every tool input expects.
\item \emph{orientation}: the gripper's pose is reported as two robot-frame unit vectors, \emph{approach} (the direction the fingers reach) and \emph{opening} (the fingertip-to-fingertip axis the jaws open/close along).
\end{itemize}
Robot Z position carries a fixed height offset, so a canonical top-down gripper reads $z \approx 1.0$, not 0; reason about position deltas.
\end{tooldoc}

\subsubsection{Observation Tools}

\begin{tooldoc}{screenshot()}
Capture a screenshot of the robot interface.
The window shows:
\begin{itemize}
\item Left sidebar: two camera feeds.
Both are rendered from the real scene, and they refresh only after \texttt{execute\_waypoint} moves the real gripper.
The \emph{third-person view} is a fixed camera looking at the table and gripper, best for analyzing the full scene layout (objects relative to each other and the gripper).
The \emph{wrist view} is a camera on the gripper looking down between the fingers, best for the close-up gripper-object relationship (whether the gripper is well positioned for a grasp or place).
\item Right main view: a 3D point-cloud reconstruction of the scene, with robot-frame axes (X red, Y green, Z blue) and the blue target gripper mesh you command.
Best for 3D and depth understanding that the flat camera images cannot give.
\end{itemize}
Call this first, or whenever you need a fresh view without taking an action.
\end{tooldoc}

\begin{tooldoc}{hover(u, v)}
Move the mouse to normalized screenshot coordinates $(u, v)$ and show a visible crosshair overlay in the returned screenshot.
The exact $(u, v)$ point is the crosshair's center dot.
Use \texttt{hover} to inspect a spot before acting on it.
It takes no action and gives two readouts:
\begin{itemize}
\item Visual: the crosshair (a persistent DOM overlay, not the OS cursor) marks exactly where $(u, v)$ lands, confirming it is on the intended UI element or point-cloud region.
\item Contact point (robot-frame $x/y/z$, in metres), computed read-only.
On the point-cloud canvas or a camera feed it reports the cloud point under the cursor (the ray need not be pixel-exact; it hits within a small radius).
Two uses: probing the robot-frame coordinate of any point of interest, and previewing a teleport, since a point it lands on is exactly the salient point that a \texttt{gripper\_teleport\_via\_click} at the same $(u, v)$ would select.
Note this is the clicked point itself, not the resulting gripper pose (which lands $\sim$0.07~m back along the approach axis).
It reads \texttt{none} when $(u, v)$ is not over a cloud point (empty space), over the blue target gripper, or over UI chrome rather than the point-cloud canvas or a camera feed.
\end{itemize}
\end{tooldoc}

\begin{tooldoc}{gripper\_get\_pose()}
Return the current blue target gripper pose and visually show its approach axis in the screenshot.
Besides giving a readout of the exact current gripper pose, the drawn approach axis makes this the best orientation check before using \texttt{gripper\_advance\_or\_retreat}: it shows which way \texttt{f}/\texttt{b} will step.
Reports the center of the fingertips of the blue target gripper in robot-frame metres, its orientation as approach and opening unit vectors, and the open/closed state.
The visual axis guide is shown without moving the gripper.
\end{tooldoc}

\begin{tooldoc}{gripper\_show\_rotation\_gizmo()}
Show the gripper's rotation gizmo (the three rotation rings) together with its approach axis in the screenshot.
Always call this first to choose the \texttt{gripper\_rotate} axis: it gives an intuitive visual understanding of the rotation axes (the x/y/z gizmo rings).
The arrow on each rotation ring points in the $+$ direction of that rotation.
Reports the center of the fingertips of the blue target gripper in robot-frame metres, its orientation as approach and opening unit vectors, and the open/closed state.
\end{tooldoc}

\begin{tooldoc}{camera\_get\_pose()}
Return the current camera pose without taking an action.
The camera orbits a target point, reported in the robot frame (metres) like every other pose: \texttt{target} is the world point the camera looks at, and \texttt{distance} is the camera-to-target separation in metres.
\texttt{azimuth} and \texttt{elevation} give the viewing direction onto that target: azimuth is the horizontal orbit angle in degrees; elevation is 0 at horizontal (level with the target) and 90 at top-down, but orbiting keeps it within $[0, 85]$ so it never reaches the degenerate top-down pole (where azimuth would collapse to 0).
The target is reported because pan and off-center zoom shift the scene center.
Use these values to reason about the current framing before adjusting it with other camera tools.
\end{tooldoc}

\subsubsection{Action Tools}

\begin{tooldoc}{gripper\_teleport\_via\_click(u, v)}
Fast-move the blue target gripper near a clicked salient point; returns a screenshot.
Best for quickly setting a subtask's starting position; refine afterward with \texttt{gripper\_drag}, \texttt{gripper\_translate}, \texttt{gripper\_rotate}, or \texttt{gripper\_toggle}.
Where to click:
\begin{itemize}
\item The left-panel 2D camera views are best when the target is clearly visible in either camera feed.
\item The right-panel 3D point-cloud view is best when locating the target takes some 3D reasoning about depth.
\item Click the object's center, not its edge, so the gripper lands centered over it for a reliable grasp or place.
\end{itemize}
The gripper lands $\sim$0.07~m back from the clicked point along its current approach axis (opposite the reach direction), not on the point itself, so a short forward move or descent then reaches it.
For example, if the gripper currently points straight down, this puts it $\sim$0.07~m directly above the click.
Translation only: the gripper's orientation is preserved.
The readout reports the clicked point as the ``selected salient point'' and the resulting gripper pose separately; check both, plus the screenshot.
Clicking the blue target gripper itself does nothing here; that is reserved for \texttt{gripper\_drag}.
\end{tooldoc}

\begin{tooldoc}{gripper\_drag(u1, v1, u2, v2, [constraint], [steps])}
Drag the blue target gripper from $(u_1, v_1)$ to $(u_2, v_2)$; returns a screenshot.
The blue target gripper is the target pose you command; this tool moves it.
The gray point-cloud gripper is actual sensor data and is not a draggable mesh.
The start point $(u_1, v_1)$ must ray-hit the solid blue target gripper mesh (a finger, wrist, or body), not the empty gap between the fingers or the gray gripper; a miss is refused with a no-op.
A free drag moves the gripper with the cursor in the view plane, $\sim$0.0006~m per CSS pixel of mouse movement, before constraints.
The optional \texttt{constraint} picks the movement mode; \texttt{x}/\texttt{y}/\texttt{z} lock motion to one robot-frame axis (the fixed robot X/Y/Z, not the gripper-attached gizmo axes that \texttt{gripper\_rotate} spins about) and are exact at any camera angle, so trust the pose readout, not the picture:
\begin{itemize}
\item \texttt{free} (default): unconstrained, follows the cursor in the camera view plane (camera-dependent).
\item \texttt{x}: lock to robot X only.
\item \texttt{y}: lock to robot Y only.
\item \texttt{z}: lock to robot Z (vertical).
\end{itemize}
Tip: \texttt{hover} first to confirm $(u_1, v_1)$ lands on the gripper, then use small single-axis drags.
The screenshot shows start and end markers, the end crosshair, the arrow path, and the updated pose.
The optional \texttt{steps} sets the number of mouse-move interpolation steps (default 25).
\end{tooldoc}

\begin{tooldoc}{gripper\_translate([d\_x], [d\_y], [d\_z])}
Translate the blue target gripper by a metric offset along the robot-frame axes, in metres; returns a screenshot.
You specify the offset directly, with no start point, mesh hit, or mouse drag needed, so it is the exact, camera-independent way to shift the gripper along the world axes.
This is the keyed equivalent of \texttt{gripper\_drag}'s x/y/z axis locks without the cursor imprecision; the orientation is unchanged.
Axes are the fixed robot X/Y/Z (the colored arrows), not the gripper-attached gizmo axes that \texttt{gripper\_rotate} spins about, so the offsets are exact at any camera angle: \texttt{d\_x} moves along robot X, \texttt{d\_y} along robot Y, and \texttt{d\_z} along robot Z (vertical: $+z$ lifts, $-z$ lowers).
Set one offset for a pure single-axis move, or several to move diagonally; omitted offsets default to 0 and at least one must be nonzero.
Each offset must be within $\pm0.1$~m to keep moves gradual; larger requests are rejected, so call repeatedly to translate further.
The returned text reports the actual pose delta after the move.
\end{tooldoc}

\begin{tooldoc}{gripper\_advance\_or\_retreat(direction, [steps])}
Move the blue target gripper along its approach axis; returns a screenshot.
Direction \texttt{f} steps along the approach axis (toward where the fingers point); \texttt{b} steps opposite it (retreat, back out).
In the robot frame the direction follows the current orientation: at a top-down pose, \texttt{f} is $-Z$ (down) and \texttt{b} is $+Z$ (lift), while on a side grasp the same steps are horizontal.
If the current orientation makes the direction ambiguous, call \texttt{gripper\_get\_pose} first to see the approach axis drawn in the screenshot.
One step is about 0.006~m along the approach axis; use 1--3 steps for small corrections and larger counts for coarse approach or retreat (default 1).
The returned text reports the actual pose delta.
\end{tooldoc}

\begin{tooldoc}{gripper\_rotate(axis, angle)}
Rotate the blue target gripper about one gizmo axis (\texttt{x}, \texttt{y}, or \texttt{z}) by a signed angle in degrees; returns a screenshot.
Important: call \texttt{gripper\_show\_rotation\_gizmo} before every rotation and pick the axis and direction from the gizmo you see; do not guess.
The x/y/z here are the gripper's own gizmo axes, not the world/robot X/Y/Z (the colored arrows); they line up only at the top-down pose and move with the gripper after any tilt, so the same letter points a different way once the gripper is tilted.
The gripper has three mutually perpendicular gizmo axes: the approach axis (where the fingers reach), the opening axis (the finger line), and a third axis perpendicular to both.
Each rotation spins about one, leaving it fixed and re-aiming the other two (the arrow on each ring shows the $+$ direction):
\begin{itemize}
\item \texttt{x}: about the third axis; both the approach and opening axes re-aim.
\item \texttt{y}: about the opening axis; opening stays fixed, approach re-aims.
\item \texttt{z}: about the approach axis; approach stays fixed, opening re-aims (twist in place).
\end{itemize}
The angle must be within $[-90, 90]$ degrees; prefer small steps (10--45 degrees) so you can observe the intermediate result and adjust before overshooting.
The returned text reports the resulting approach and opening axes; the rotation is in place (position unchanged).
If you are unsure about the axis or direction, rotate by a small amount and observe the outcome.
\end{tooldoc}

\begin{tooldoc}{gripper\_toggle([state])}
Open or close the blue target gripper; returns a screenshot.
Wraps the UI \texttt{g} key, which swaps the gripper mesh between open and closed.
Pass \texttt{state} to drive a known result without checking the current state first: \texttt{toggle} (default) flips the current state, \texttt{open} ensures open (no-op if already open), and \texttt{closed} ensures closed (no-op if already closed).
The returned message and screenshot report the resulting open/closed state, read back after the swap.
\end{tooldoc}

\begin{tooldoc}{gripper\_reset()}
Reset the blue target gripper back onto the real gripper's current pose (position, orientation, and open/closed state); returns a screenshot.
Use this to undo experimental edits: freely try \texttt{gripper\_drag}, \texttt{gripper\_advance\_or\_retreat}, \texttt{gripper\_rotate}, or \texttt{gripper\_toggle} to plan a move, see how it looks, then reset to the real pose before committing the move you actually want.
This only moves the blue target gripper; it does not move the real robot and records no waypoint.
\end{tooldoc}

\subsubsection{Camera Tools}

\begin{tooldoc}{camera\_orbit\_via\_key(d\_azimuth, d\_elevation)}
Orbit the camera around the scene target by an angular delta, in degrees; returns a screenshot.
You specify the two orbit angles directly, with no start point or open-space click needed, so it is the simplest way to reframe.
Both deltas are relative to the current pose (read \texttt{camera\_get\_pose} for the absolute angles); the distance and look-at target are unchanged, as this only rotates the viewpoint.
Positive \texttt{d\_azimuth} swings the camera counterclockwise seen from above, negative swings it clockwise; at the default view, azimuth is 90 with the camera at $Y=0$ looking toward $-X$, and positive deltas swing it toward $+Y$, reaching a view toward $-Y$ at azimuth 180.
Positive \texttt{d\_elevation} raises the camera toward a top-down view (elevation toward 90); negative lowers it toward a front or side horizontal view (elevation toward 0).
Each delta must be within $\pm45$ degrees to keep camera moves gradual; larger requests are rejected, so call repeatedly to orbit further.
Elevation is clamped to $[0, 85]$ (the horizon is allowed; the exact top-down pose is degenerate and kept just out of reach); azimuth wraps at 360.
Returns the updated camera pose.
\end{tooldoc}

\begin{tooldoc}{camera\_pan\_via\_key(d\_right, d\_up)}
Pan the camera (shift its look-at point) by a metric offset, in robot-frame metres; returns a screenshot.
You specify the shift directly, with no start point or drag needed, so it is the simplest way to re-center the view without rotating.
The viewing angle (azimuth/elevation) and distance are unchanged: the camera and its target slide together by the same amount, so only the framing moves.
Positive \texttt{d\_right} pans right, sliding the look-at point toward screen-right so you see more of what was off the right edge (scene content appears to slide left); negative pans left.
Positive \texttt{d\_up} pans up along world-Z (robot $+z$) so you see more above (scene content appears to slide down); negative pans down.
Each offset must be within $\pm0.2$~m to keep moves gradual; larger requests are rejected, so call repeatedly to pan further.
Returns the updated camera pose, whose target shows the new framing center.
\end{tooldoc}

\begin{tooldoc}{camera\_zoom(u, v, [steps])}
Zoom the camera toward normalized screenshot coordinates; returns a screenshot.
The browser zooms toward the point under the mouse; $(u, v)$ should be inside the point-cloud view.
Positive \texttt{steps} zoom in and negative zoom out (default 1); internally this sends scroll-wheel events to the browser.
Each step changes the camera-to-target distance by a constant $\sim$0.03~m, regardless of the current zoom level or direction, so the step count is roughly the desired distance change divided by 0.03~m.
An off-center zoom also shifts composition toward that point; use this deliberately to reframe while zooming.
Returns the updated camera pose.
\end{tooldoc}

\begin{tooldoc}{camera\_reset()}
Restore the initial camera view; returns a screenshot and the updated camera pose.
Strategy: the default view is framed so most operations are viable from it, so prefer working there and return to it with this tool after exploring.
Reach for \texttt{camera\_orbit\_via\_key} or \texttt{camera\_pan\_via\_key} only when the default view is insufficient, e.g., to resolve an occlusion or to judge the gripper-to-object relationship (depth, alignment, clearance) from another angle, then reset here to re-establish a known frame.
\end{tooldoc}

\subsubsection{Execution Tools}

\begin{tooldoc}{execute\_waypoint()}
Execute the current gripper target and record one waypoint; returns a screenshot.
This asks the robot controller to move the real gripper to the current target pose: position, orientation, and final open/closed gripper state.
Use this after meaningful target-setting operations such as \texttt{gripper\_teleport\_via\_click}, \texttt{gripper\_drag}, \texttt{gripper\_advance\_or\_retreat}, \texttt{gripper\_rotate}, or \texttt{gripper\_toggle}.
Execute a \texttt{gripper\_toggle} on its own; do not combine it with a pose change in the same waypoint, since moving and opening/closing the gripper at once is unreliable.
Execution uses the final target as the next base state for later operations.
The controller interpolates between the current pose and the target pose; it does not replay the exact UI trajectory used to place the target.
For example, if you toggled the gripper twice before executing, it only tries to match the final open/closed state, without intermediate toggles.
If this waypoint achieves the task goal, the result reports \texttt{TASK SUCCEEDED} and that the connection is about to close.
When you see that, stop: the episode is done and no further tools should be called.
\end{tooldoc}

\begin{tooldoc}{end\_episode()}
Save and end the current episode and reset or advance the task; returns a screenshot.
Call this only after the final waypoint for the episode has been executed.
\end{tooldoc}

\endgroup

\subsection{Real-World Tools}
\label{sec:appendix-real-tools}

% Source: https://github.com/hengyuan-hu/via-dev/blob/main2d/sphinx/ui/mcp_server.py
% Git blob: 6259060128c4948525a3293fa2b8535dff37c6f9
% Descriptions use the Stretch view names (head and wrist), with wrist depth enabled.
% Runtime placeholders are expanded as in _patch_tools_for_views.

The real-world interface returns named camera views with a virtual target gripper overlay and a JSON readout of the real and target gripper poses.
\cref{tab:real-tools} summarizes the tools for the Stretch 4 robot.
The descriptions below use its \texttt{head} and \texttt{wrist} views, with wrist depth enabled.
When wrist depth is disabled, \texttt{hover} still shows a crosshair in the wrist view but provides no depth readout there, and \texttt{show\_approach\_axis} is not offered.

\suppressfloats[t]
\begin{table}[ht]
\centering
\renewcommand{\arraystretch}{1.1}
\begin{tabular}{@{}l >{\raggedright\arraybackslash}p{10.9cm}@{}}
\toprule
\textbf{Category} & \textbf{Tools} \\
\midrule
Observation & \texttt{observe}, \texttt{hover}, \texttt{show\_approach\_axis} \\
\addlinespace[3pt]
Action & \texttt{ee\_teleport\_via\_click}, \texttt{ee\_robot\_centric\_translate}, \texttt{ee\_wrist\_centric\_translate}, \texttt{ee\_wrist\_centric\_rotate}, \texttt{gripper\_set}, \texttt{reset\_target} \\
\addlinespace[3pt]
Navigation & \texttt{nav\_translate}, \texttt{nav\_rotate} \\
\addlinespace[3pt]
Execution & \texttt{execute\_waypoint}, \texttt{end\_episode} \\
\bottomrule
\end{tabular}
\caption{\textbf{MCP tools for real-world experiments.}
The agent observes named camera views, stages end-effector or mobile-base waypoints, and moves the robot through \texttt{execute\_waypoint}.
Most tools return updated views and a \texttt{proprio} readout; the observation probes and message-only results are described below.}
\label{tab:real-tools}
\end{table}

Positions and translation offsets are in metres; rotation angles are in degrees.
The robot axes are $+x$ forward, $+y$ left, and $+z$ up, relative to the current real base pose.
The wrist-centric tools instead use the target gripper's axes, defined in their descriptions.
Optional numeric arguments default to zero.
A waypoint stages one of three kinds of command: base navigation, end-effector pose, or gripper opening/closing.
These kinds cannot be combined in a single waypoint; \texttt{execute\_waypoint} executes the staged command and \texttt{reset\_target} discards it.

\begingroup
\setlist[itemize]{leftmargin=1.5em, itemsep=1pt, topsep=2pt, parsep=0pt, beginpenalty=10000}

\subsubsection{Observation Tools}

\begin{tooldoc}{observe()}
Return the latest observation of the world, as these blocks in order:

\begin{itemize}
\item for each view listed below, a \texttt{view: <name>} label followed by that view's image;
\item a JSON dict with one key, \texttt{proprio}, containing:
\begin{itemize}
\item \texttt{mode}: the staged waypoint, one of \texttt{none}, \texttt{nav}, \texttt{gripper}, \texttt{pose};
\item \texttt{real} and \texttt{target}: the real and the target gripper pose, each with \texttt{position\_m}, the \texttt{approach} and \texttt{opening} unit vectors as x/y/z on the robot axes, and the jaw state \texttt{gripper\_open}.
\end{itemize}
\end{itemize}

The views, in the order returned:
\begin{itemize}
\item \texttt{head}: a camera mounted on top of the Stretch 4 robot; it moves with the base but is fixed relative to it.
A red virtual target gripper is drawn into this view.
It starts at the real gripper's pose; tools move it to stage a waypoint, and it is updated after each relevant tool call so you can see the staged effect before executing.
After \texttt{execute\_waypoint} or \texttt{reset\_target} it snaps back onto the real gripper.
\item \texttt{wrist}: a camera mounted on the wrist (the last joint connecting the gripper to the rest of the robot), so this view moves whenever the gripper moves.
The target is drawn here as a red \texttt{|----.----|} glyph instead of the full gripper shape, so it never hides the object being grasped: the two end ticks mark where the target fingertips would be, the line between them spans the jaw opening, and the center dot marks the closure point where the jaws meet.
Use this view for fine alignment right before grasping.
\end{itemize}
\end{tooldoc}

\begin{tooldoc}{hover(view, u, v)}
Show a crosshair at (u, v) in the named view and report the scene point under it (when available).

\emph{Arguments.}
\begin{itemize}
\item \texttt{view}: \texttt{head} or \texttt{wrist};
\item \texttt{u}: fraction across that view (0.0 left $\rightarrow$ 1.0 right);
\item \texttt{v}: fraction down (0.0 top $\rightarrow$ 1.0 bottom).
\end{itemize}

\emph{Returns.}
\begin{itemize}
\item The chosen view with a crosshair overlay at (u, v).
\item If the view is depth-augmented, a JSON dict with \texttt{offset\_m}, the hovered scene point in metres relative to the center of the real gripper (not the staged target), as x/y/z on the robot axes; or, when no scene point resolves under the crosshair, a message saying so instead (an \texttt{ee\_teleport\_via\_click} there would be rejected).
\end{itemize}

\emph{Tips.}
\begin{itemize}
\item Use it to check whether (u, v) lands on the desired part before \texttt{ee\_teleport\_via\_click}.
\item The offset is computed from depth, which can be noisy (especially on the real robot).
You should always trust the image over the offset readout.
\end{itemize}
\end{tooldoc}

\begin{tooldoc}{show\_approach\_axis()}
Show where the target gripper's approach axis meets the scene, as a one-shot X in the wrist view.

The approach axis is where the center of the gripper is pointing at;
it is also the line \texttt{ee\_wrist\_centric\_translate}'s \texttt{dx\_m} moves along.

\emph{Returns.}
\begin{itemize}
\item The wrist view with a light-blue X where the approach axis first meets a surface.
\item Only when the X is absent, a JSON dict with \texttt{message} saying why: no wrist depth is available, or the axis meets no surface within 2 m of the target.
\item A message-only rejection while the current waypoint was placed by \texttt{ee\_teleport\_via\_click}: that pose came from the scene cloud, and this guide is wrist-only.
In that case, \texttt{execute\_waypoint} or \texttt{reset\_target} first.
\end{itemize}

\emph{Tips.}
\begin{itemize}
\item This tool is useful to check where the gripper is pointing before advancing the gripper for grasp.
\item The X comes from wrist depth, which can be noisy on the real robot: always trust the image over the X.
\end{itemize}
\end{tooldoc}

\subsubsection{Action Tools}

\begin{tooldoc}{ee\_teleport\_via\_click(u, v)}
Teleport the target gripper to a scene point clicked in the head view: it lands 0.07 m back from the clicked point along its current approach axis, orientation preserved.

\emph{Arguments.}
\begin{itemize}
\item \texttt{u}: fraction across the head view (0.0 left $\rightarrow$ 1.0 right);
\item \texttt{v}: fraction down (0.0 top $\rightarrow$ 1.0 bottom).
\end{itemize}

\emph{Returns.}
\begin{itemize}
\item The views with the target gripper updated to the staged pose; the head view also marks the clicked point with a red dot.
\item A JSON dict with \texttt{offset\_m}, the clicked scene point in metres relative to the center of the real gripper (not the staged target), as x/y/z on the robot axes, and the usual \texttt{proprio}.
\item If a nav or gripper waypoint is staged, or no scene point resolves under the click, a single rejection message instead; nothing is staged.
\end{itemize}

\emph{Tips.}
\begin{itemize}
\item When uncertain, use \texttt{hover} to check that the click lands on the desired part.
\item Rotate the gripper first if the grasp needs a different orientation.
\item The landing depth comes from the scene cloud, which is noisy on the real robot.
Use this for coarse repositioning, then refine with the other tools.
\end{itemize}
\end{tooldoc}

\begin{tooldoc}{ee\_robot\_centric\_translate([dx\_m], [dy\_m], [dz\_m])}
Translate the target gripper along the robot-centric axes.

\emph{Arguments.}
\begin{itemize}
\item \texttt{dx\_m}: metres, forward (negative = backward), within $\pm0.5$;
\item \texttt{dy\_m}: metres, left (negative = right), within $\pm0.5$;
\item \texttt{dz\_m}: metres, up (negative = down), within $\pm0.5$.
\end{itemize}

The cumulative staged gripper travel from the real pose (across pose edits since the last execute) is capped at 1.5 m per waypoint.

\emph{Returns.}
\begin{itemize}
\item The views with the target gripper updated to the staged pose, and the usual \texttt{proprio}.
\item If a cap is exceeded, per call or cumulative, or a nav or gripper waypoint is staged, a single rejection message instead; nothing is staged.
\end{itemize}
\end{tooldoc}

\begin{tooldoc}{ee\_wrist\_centric\_translate([dx\_m], [dy\_m], [dz\_m])}
Translate the target gripper on the wrist-centric axes.

The axes are the target gripper's, not the real gripper's.
A staged rotation will change those axes so we always reason with respect to the target gripper.

\emph{Arguments.}
\begin{itemize}
\item \texttt{dx\_m}: metres along x, the approach axis (positive = toward where the fingers point);
\item \texttt{dy\_m}: metres along y, the jaw-opening line (positive = the wrist view's left);
\item \texttt{dz\_m}: metres along z, perpendicular to both (positive = the wrist view's up).
\end{itemize}

\texttt{dx\_m}, \texttt{dy\_m} and \texttt{dz\_m} are within $\pm0.2$ m.

\emph{Returns.}
\begin{itemize}
\item The views with the target gripper updated to the staged pose, and the usual \texttt{proprio}.
\item The wrist view also carries a light-blue X where the approach axis of the updated target gripper first meets a surface, as \texttt{show\_approach\_axis} would draw it; it is omitted after any \texttt{ee\_teleport\_via\_click} since the last \texttt{execute\_waypoint} or \texttt{reset\_target}, even if wrist edits followed.
When the X is absent for another reason, \texttt{message} says why.
\item If a cap is exceeded, or a nav or gripper waypoint is staged, a single rejection message instead; nothing is staged.
\end{itemize}
\end{tooldoc}

\begin{tooldoc}{ee\_wrist\_centric\_rotate([roll\_deg], [tilt\_up\_deg], [tilt\_left\_deg])}
Rotate the target gripper on the wrist-centric axes.

The axes are the target gripper's, not the real gripper's.
A staged rotation will change those axes so we always reason with respect to the target gripper.

\emph{Arguments.}
\begin{itemize}
\item \texttt{roll\_deg}: degrees about x, the approach axis (positive = counter-clockwise in the wrist view); the jaws spin in place;
\item \texttt{tilt\_up\_deg}: degrees tipping x toward z (positive = toward the wrist view's up);
\item \texttt{tilt\_left\_deg}: degrees tipping x toward y (positive = toward the wrist view's left).
\end{itemize}

\texttt{tilt\_up\_deg} and \texttt{tilt\_left\_deg} pivot at the wrist mount behind the jaws, so the closure point sweeps an arc.

Measured Stretch wrist travel from neutral (one physical joint at a time):
\begin{itemize}
\item Physical roll: -61.9 to +65 deg; near neutral, positive \texttt{roll\_deg} moves physical roll in the negative direction.
\item Physical pitch: -61.8 to +75.0 deg; near neutral, positive \texttt{tilt\_up\_deg} moves physical pitch in the negative direction.
\item Physical yaw: -62.8 to +89.9 deg; near neutral, positive \texttt{tilt\_left\_deg} moves physical yaw in the positive direction.
\end{itemize}
These are successfully tested excursions, not mechanical endpoints or guaranteed ranges after combined rotations.

Each argument is limited to $\pm45$ deg per call.
Angles are relative to the current target pose; repeated calls accumulate rotation.
The measured joint ranges above are total angles from neutral, not per-call allowances; available travel depends on the current pose.

\emph{Returns.}
\begin{itemize}
\item The views with the target gripper updated to the staged pose, and the usual \texttt{proprio}.
\item The wrist view also carries a light-blue X where the approach axis of the updated target gripper first meets a surface, as \texttt{show\_approach\_axis} would draw it; it is omitted after any \texttt{ee\_teleport\_via\_click} since the last \texttt{execute\_waypoint} or \texttt{reset\_target}, even if wrist edits followed.
When the X is absent for another reason, \texttt{message} says why.
\item If a cap is exceeded, or a nav or gripper waypoint is staged, a single rejection message instead; nothing is staged.
\end{itemize}

\emph{Tips.}
\begin{itemize}
\item Combined angles apply in order roll, \texttt{tilt\_up}, \texttt{tilt\_left}.
\item Split into multiple calls if you want a different order.
\end{itemize}
\end{tooldoc}

\begin{tooldoc}{gripper\_set(state)}
Stage the jaw state of the target gripper.

A gripper-only waypoint: it cannot combine with a staged nav or pose edit and needs its own \texttt{execute\_waypoint}.

\emph{Arguments.}
\begin{itemize}
\item \texttt{state}: \texttt{open} or \texttt{closed}.
\end{itemize}

\emph{Returns.}
\begin{itemize}
\item The views with the target gripper's jaws updated to the staged state, and the usual \texttt{proprio}; the real jaws move only on \texttt{execute\_waypoint}.
\item If the target is already staged that way, or a nav or pose edit is staged, a single message explaining the rejection; nothing changes.
\end{itemize}
\end{tooldoc}

\begin{tooldoc}{reset\_target()}
Discard all staged edits on the target gripper.
The target gripper snaps back onto the real gripper's pose and jaw state, and the waypoint mode resets to none.

\emph{Returns.}
\begin{itemize}
\item The views with the target gripper re-synced, and the usual \texttt{proprio}.
\item If nothing is staged, or on an error, a single message instead.
\end{itemize}
\end{tooldoc}

\subsubsection{Navigation Tools}

\begin{tooldoc}{nav\_translate([dx\_m], [dy\_m])}
Stage a mobile-base translation on the robot-centric axes (+x forward, +y left).

\emph{Arguments.}
\begin{itemize}
\item \texttt{dx\_m}: metres forward along x-axis (negative = backward), within $\pm0.5$;
\item \texttt{dy\_m}: metres left along y-axis (negative = right), within $\pm0.5$.
\end{itemize}

dx/dy follow the staged heading: after a staged \texttt{nav\_rotate}, forward/left are along the rotated axes.
The cumulative staged base translation (across nav calls since the last execute) is capped at 0.5 m per waypoint.

\emph{Returns.}
\begin{itemize}
\item The head view with the target gripper updated to the staged pose (it moves rigidly with the base and may leave the viewport), and a top-down \texttt{nav\_map} view in place of the wrist view (staged base motion cannot move the wrist camera): range rings every 0.1 m out to 0.5 m, the gray footprint is the real base pinned at center, the red footprint with the nose wedge is the staged base.
\item A JSON dict with \texttt{staged\_base\_delta} (\{\texttt{dx\_m}, \texttt{dy\_m}, \texttt{dyaw\_deg}\}) that captures the cumulative staged base displacement (if there have been multiple \texttt{nav\_\{translate/rotate\}} calls), given on the robot's current real axes, not the staged axes, and the usual \texttt{proprio}.
\item If a cap is exceeded, per call or cumulative, or a gripper or pose edit is staged, a single message explaining the rejection instead; nothing is staged.
\end{itemize}

\emph{Tips.}
\begin{itemize}
\item This tool is useful to move closer to a target location to perform subsequent manipulation.
\end{itemize}
\end{tooldoc}

\begin{tooldoc}{nav\_rotate(angle\_deg)}
Stage a mobile-base rotation in place, about the +z (up) axis.

\emph{Arguments.}
\begin{itemize}
\item \texttt{angle\_deg}: degrees about +z (positive = counter-clockwise = turn left; negative = turn right), within $\pm90$.
\end{itemize}

The cumulative staged base rotation (across nav calls since the last execute) is capped at 90 deg per waypoint.

\emph{Returns.}
\begin{itemize}
\item The head view with the target gripper updated to the staged pose (it swings along an arc around the base center and may leave the viewport), and a top-down \texttt{nav\_map} view in place of the wrist view (staged base motion cannot move the wrist camera): range rings every 0.1 m out to 0.5 m, the gray footprint is the real base pinned at center, the red footprint with the nose wedge shows the staged heading.
\item A JSON dict with \texttt{staged\_base\_delta} (\{\texttt{dx\_m}, \texttt{dy\_m}, \texttt{dyaw\_deg}\}) that captures the cumulative staged base displacement (if there have been multiple \texttt{nav\_\{translate/rotate\}} calls), given on the robot's current real axes, not the staged axes, and the usual \texttt{proprio}.
\item If a cap is exceeded, per call or cumulative, or a gripper or pose edit is staged, a single message explaining the rejection instead; nothing is staged.
\end{itemize}

\emph{Tips.}
\begin{itemize}
\item \texttt{nav\_rotate} turns the robot-centric axes with the base: after a rotation (staged or executed), +x and +y point in the new directions, so a later \texttt{nav\_translate} follows the rotated heading.
\item The head camera also turns with this tool, so this tool is helpful for searching for and locating the point of interest.
\end{itemize}
\end{tooldoc}

\subsubsection{Execution Tools}

\begin{tooldoc}{execute\_waypoint()}
Execute the staged waypoint on the actual robot; nothing moves until this call; it blocks until execution finishes.

\emph{Returns.}
\begin{itemize}
\item The views and \texttt{proprio} as usual, target snapped onto the realized pose.
\item \texttt{ordered\_vs\_realized} \{mode, ordered, realized\}: commanded vs measured.
Per mode:
\begin{itemize}
\item nav: base displacements \{\texttt{dx\_m}, \texttt{dy\_m}, \texttt{dyaw\_deg}\} on the pre-execution heading, realized from odometry;
\item pose: gripper poses in \texttt{proprio}'s format, both on the robot axes after execution; if the base moved to reach the target, \texttt{ordered} differs from the previously staged \texttt{target} by that base motion;
\item gripper: the jaw states.
\end{itemize}
\item \texttt{message}, when present: the execution was cut short (the 30 s cap or a stall; the views show where the robot stopped), or the task-success directive.
\item If nothing is staged, a single message instead; nothing executes.
\end{itemize}
\end{tooldoc}

\begin{tooldoc}{end\_episode()}
Save and end the current episode.

\emph{Returns.}
\begin{itemize}
\item A confirmation or error (if any) message.
\end{itemize}
\end{tooldoc}

\endgroup

\clearpage
\section{Tool Call Efficiency and Estimated Costs}
\label{sec:appendix-efficiency}
\suppressfloats[t]

% Values and per-episode provenance: graphics/data/efficiency.json.
% Models, tasks, and successful seeds match Table 1 (graphics/data/scaling.json).
\begin{table}[ht]
\centering
\setlength{\tabcolsep}{4pt}
\begin{tabular}{l cccccc}
\toprule
& \multicolumn{3}{c}{Claude Code} & \multicolumn{3}{c}{Codex} \\
\cmidrule(lr){2-4} \cmidrule(lr){5-7}
& Medium & Large & Extra & Medium & Large & Extra \\
Task & Sonnet 5 & Opus 4.8 & Fable 5 & Terra 5.6 & Sol 5.6 & Astra 6 \\
\midrule
Turn on stove & 84 & 36 & 39 & 46 & 48 & \textbf{16} \\
Open drawer & 117 & 47 & 38 & 91 & 33 & \textbf{19} \\
Put bowl on plate & 174 & 59 & 53 & 72 & 52 & \textbf{34} \\
Square & 81 & 92 & 69 & -- & 87 & \textbf{42} \\
Rainbow & 271 & 159 & 143 & 175 & 208 & \textbf{136} \\
T-block & -- & 81 & 89 & 162 & 109 & \textbf{59} \\
\midrule
Overall & 146 & 79 & 72 & 109 & 90 & \textbf{51} \\
\bottomrule
\end{tabular}
\caption{\textbf{The average number of tool calls per successful episode.}
Mean number of MCP tool calls over successful episodes for the default-prompt evaluations in \cref{tab:scaling-results}.
`--' indicates that no episode succeeded.
Bold marks the fewest tool calls in each row.
The Overall row averages task means over tasks with at least one successful episode.
}
\label{tab:tool-calls}
\end{table}

\begin{table}[ht]
\centering
\setlength{\tabcolsep}{4pt}
\begin{tabular}{l cccccc}
\toprule
& \multicolumn{3}{c}{Claude Code} & \multicolumn{3}{c}{Codex} \\
\cmidrule(lr){2-4} \cmidrule(lr){5-7}
& Medium & Large & Extra & Medium & Large & Extra \\
Task & Sonnet 5 & Opus 4.8 & Fable 5 & Terra 5.6 & Sol 5.6 & Astra 6 \\
\midrule
Turn on stove & \$\,2.8 & \$\,4.1 & \$\,8.9 & \$\,1.1 & \$\,2.5 & \textbf{\$\,1.0} \\
Open drawer & \$\,4.8 & \$\,7.4 & \$\,7.5 & \$\,2.8 & \textbf{\$\,1.0} & \$\,1.3 \\
Put bowl on plate & \$\,5.8 & \$\,7.5 & \$\,11.8 & \$\,1.9 & \textbf{\$\,1.7} & \$\,2.5 \\
Square & \textbf{\$\,2.8} & \$\,12.5 & \$\,14.1 & -- & \$\,3.5 & \textbf{\$\,2.8} \\
Rainbow & \$\,10.2 & \$\,26.0 & \$\,36.1 & \textbf{\$\,1.6} & \$\,4.5 & \$\,7.6 \\
T-block & -- & \$\,9.7 & \$\,19.2 & \$\,4.9 & \$\,3.8 & \textbf{\$\,3.7} \\
\midrule
Overall & \$\,5.3 & \$\,11.2 & \$\,16.3 & \textbf{\$\,2.4} & \$\,2.8 & \$\,3.2 \\
\bottomrule
\end{tabular}
\caption{\textbf{The average API cost per successful episode.}
Estimated mean API cost in USD over successful episodes for the default-prompt evaluations in \cref{tab:scaling-results}.
Costs are estimated from recorded token usage using model-specific API prices.
`--' indicates that no episode succeeded.
Bold marks the lowest cost in each row, including ties at the reported precision.
The Overall row averages task means equally over tasks with at least one successful episode.
}
\label{tab:cost}
\end{table}

We report the mean number of MCP tool calls in \cref{tab:tool-calls} and the average API cost in \cref{tab:cost}, both over successful episodes.
We omit failed episodes because the agents often keep trying until the episode time limit, which would significantly skew the numbers and make them highly correlated with the success rates.

\clearpage
\section{Performance Numbers of the Baseline}
\label{sec:appendix-scaling}
\suppressfloats[t]

\begin{table}[ht]
\centering
\setlength{\tabcolsep}{4pt}
\begin{tabular}{l cc ccc}
\toprule
& \multicolumn{2}{c}{Claude Code} & \multicolumn{3}{c}{Codex} \\
\cmidrule(lr){2-3} \cmidrule(lr){4-6}
Tier & Large & Extra & Medium & Large & Extra \\
Task & Opus 4.8 & Fable 5 & Terra 5.6 & Sol 5.6 & Astra 6 \\
\midrule
Turn on stove & 0\% & 50\% & 0\% & 10\% & 100\% \\
Open drawer & 10\% & 20\% & 10\% & 30\% & 80\% \\
Put bowl on plate & 0\% & 20\% & 10\% & 40\% & 100\% \\
Square & 10\% & 20\% & 10\% & 20\% & 80\% \\
Rainbow & 20\% & 20\% & 10\% & 50\% & 100\% \\
T-block & 0\% & 10\% & 0\% & 0\% & 100\% \\
\midrule
Overall & 6.7\% & 23.3\% & 6.7\% & 25.0\% & 93.3\% \\
\bottomrule
\end{tabular}
\caption{\textbf{Results for the naive baseline.}
Success rates over ten seeds per task.}
\label{tab:scaling-baseline}
\end{table}

We report per-task success rates for the naive baseline in \cref{tab:scaling-baseline}, complementing the aggregate results in \cref{fig:scaling}.
The baseline uses the same observations, agent, and controller as \via, but sends commands directly to the controller without staging or refining them in the virtual workspace.
It has only five MCP tools: \texttt{observe}, \texttt{get\_proprio}, \texttt{set\_pose}, \texttt{set\_gripper}, and \texttt{end\_episode}.
\texttt{set\_pose} and \texttt{set\_gripper} immediately invoke the controller to move the end-effector to a specified 6-DoF pose and set the gripper's open/closed state, respectively.
\texttt{observe} returns the same observation as \texttt{screenshot} in \via.
\texttt{get\_proprio} returns the robot's proprioceptive state, and \texttt{end\_episode} terminates the episode.

\clearpage
\section{Real-World Robot Platform}
\label{sec:appendix-real-robot}
\suppressfloats[t]

\begin{figure}[ht]
\centering
% Original photograph and edit provenance: graphics/assets/robot/README.md.
\includegraphics[width=0.72\linewidth]{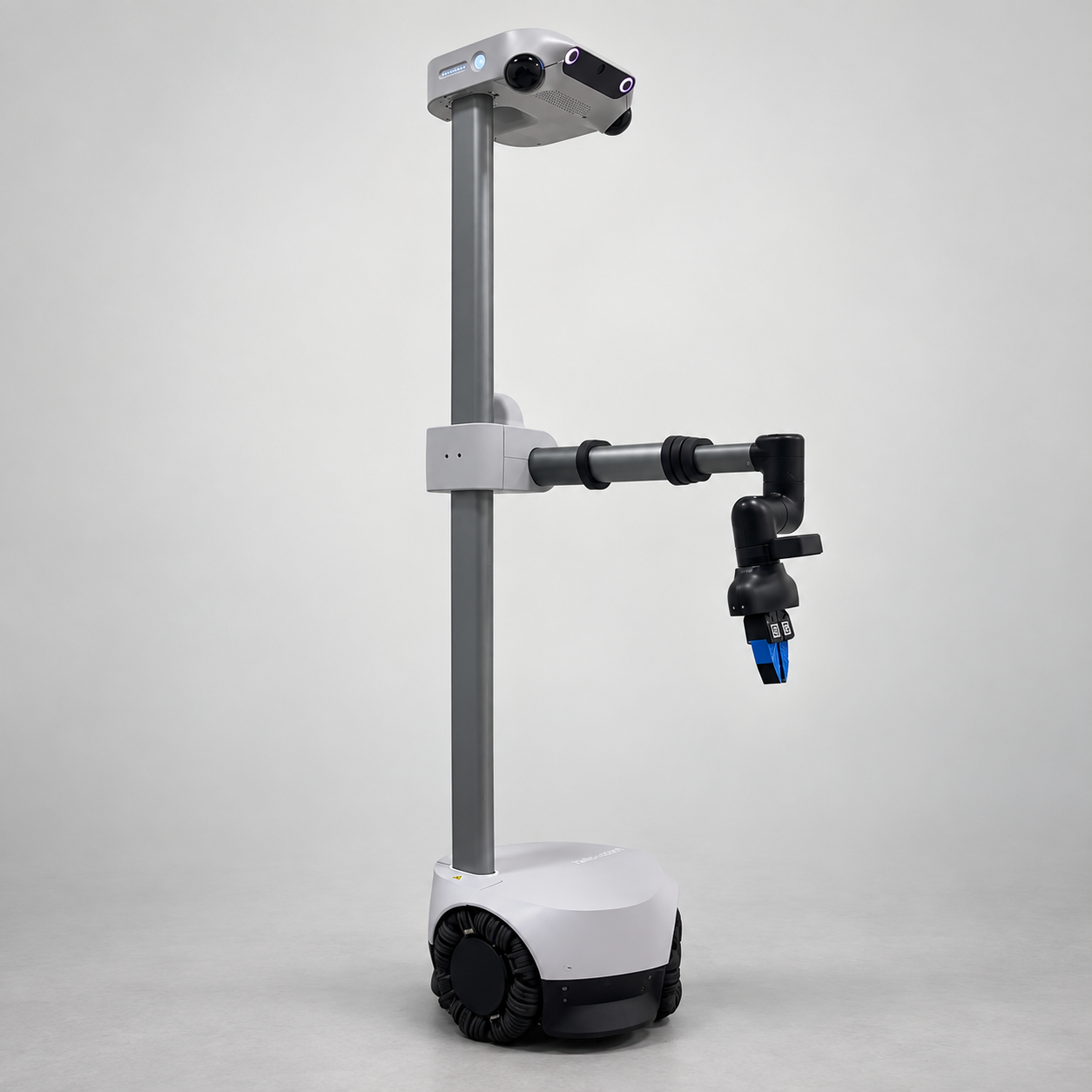}
\caption{\textbf{Real-world robot platform.}
The Stretch 4 mobile manipulator used for the five real-world tasks in \cref{sec:real-world-eval}.
}
\label{fig:real-robot-platform}
\end{figure}

Stretch 4 is a recently introduced mobile manipulation platform with a 3-DoF omnidirectional base, a 1-DoF vertical lift along the torso, a 1-DoF telescoping arm, and a 3-DoF wrist~\citep{hellorobot2026stretch4} (\cref{fig:real-robot-platform}).
We use its head camera, which is fixed relative to the base, for visual observations and its head-mounted LiDAR sensors for scene-level depth sensing.
The robot also has a wrist-mounted stereo camera, from whose images we compute depth using Fast-FoundationStereo~\citep{wen2026fastfoundationstereo}.

% Always keep the full prompts as the last appendix; add new appendices above.
\clearpage
\section{Prompts}
\label[appendix]{sec:appendix-prompts}

% Same card style for quoted prompts, with a bold title instead of monospace.
\newtcolorbox{promptdoc}[1]{
  enhanced, breakable,
  frame hidden, boxrule=0pt, arc=3pt,
  colback=gray!8,
  colbacktitle=gray!25, coltitle=black, fonttitle=\bfseries,
  title={#1},
  left=6pt, right=6pt, top=5pt, bottom=5pt,
  toptitle=3pt, bottomtitle=3pt,
  before skip=6pt, after skip=6pt,
}

\subsection{System Prompt}
\label{sec:appendix-system-prompt}

For the simulation experiments, the system prompt covers the basics of the interface (target gripper, waypoints, camera feeds, point cloud workspace) and general guidance on using it effectively.
Internal project names and file paths are redacted.

\begingroup
\setlist[itemize]{leftmargin=1.5em, itemsep=1pt, topsep=2pt, parsep=0pt}

\begin{promptdoc}{Robot Control Guide}
You operate a simulated robot arm through the MCP server to complete manipulation tasks.

\smallskip\noindent\textbf{General Information.}
You control the robot via a 3D browser GUI, similar to 3D design software or a 3D game.
The scene contains a robot arm with a gripper and a tabletop environment where you perform manipulations as instructed.

You control the robot by setting and executing a sequence of ``waypoints'':
\begin{itemize}
\item There is a blue virtual gripper in the scene, called the ``target gripper''.
You manipulate this target gripper via tools to set a ``waypoint''.
\item Repeat the following to complete a task, one waypoint at a time:
  \begin{itemize}
  \item Set: move the target gripper to a desired pose (waypoint) via the tools defined in the MCP server.
  \item Execute: A controller will bring the real gripper to the target gripper via a linear interpolation while obeying physical constraints.
  \item Check: After the waypoint is reached, the tool will return an image showing the updated UI (outcome).
  Decide the next pose based on the outcome.
  If the outcome isn't what you intended, diagnose why it failed and retry that step or any earlier one that caused it.
  \end{itemize}
\end{itemize}
For example, to pick up a cube: move the gripper above it, descend until the cube sits between the fingers, close, then lift --- each step is one waypoint that you set and then execute before moving on to the next.

\smallskip\noindent\textbf{Tips: Decomposition and Planning.}
\begin{itemize}
\item Decompose the task into \textbf{small subtasks} by breaking the goal into a short sequence of simple stages.
\item Prefer closed-loop control over open-loop.
Adjust the plan as you observe the outcome of previous waypoints.
\item Give \texttt{gripper\_toggle} its own \texttt{execute\_waypoint} cycle.
Do NOT combine it with another pose change (translate/rotate/advance...) in one waypoint.
Moving and toggling the gripper at the same time could be unpredictable.
\end{itemize}

\smallskip\noindent\textbf{Tips: Precision.}
\begin{itemize}
\item Be precise with \texttt{gripper\_teleport\_via\_click} --- it's the best way to set a good initial position.
  \begin{itemize}
  \item The clicked point lands on the gripper's approach axis, so a precise click will make future approaching easy.
  \item Use \texttt{hover} to double-check whether the predicted $(u, v)$ would land where you intend to click.
  \end{itemize}
\item Placing an object onto another object is a precision-sensitive task.
First reason about where the held object should land, and then reason about where to set the target gripper, accounting for the relation between the gripper and the held object.
\item Always check the visual from the point cloud or the camera feeds.
Do NOT solely rely on the coordinates from \texttt{hover}.
When necessary, make fine adjustments with translate, rotate, advance\_or\_retreat, and similar operations.
\end{itemize}

\smallskip\noindent\textbf{Tips: Viewing and Camera Movement.}
\begin{itemize}
\item Two camera feeds on the left.
  \begin{itemize}
  \item They provide raw camera images from 3rd-person view and wrist view.
  \item The 3rd-person view is good for understanding the scene and disambiguating objects.
  \item The wrist view is good for close-up analysis and fine adjustments.
  \item \texttt{gripper\_teleport\_via\_click} also accepts clicking on a camera feed.
  Sometimes this could be more convenient than clicking on the point cloud.
  \end{itemize}
\item The 3D point cloud on the right.
  \begin{itemize}
  \item This is the main workspace.
  It constructs the 3D scene from multiple depth cameras.
  \item The initial view (resettable via \texttt{camera\_reset}) is framed to be good for most operations.
  It is a good default so prefer working in it.
  \item A pure top-down view (elevation $\geq 80$) may not be as useful as you think, because the gripper and robot arm may block most of the workspace.
  \item Sideview (orbit by changing azimuth $\pm 30$--$45$) or frontview (elevation $\sim 10$) could be very helpful if object-gripper or object-object relation is hard to interpret in the initial default view.
  \item Move the camera gently, and reset to the initial view via \texttt{camera\_reset} to recover from a failed camera operation.
  \end{itemize}
\end{itemize}

\smallskip\noindent\textbf{Tips: Common Failures.}
\begin{itemize}
\item Grasping too high is a common failure.
Descend deep so that the fingers and the object to hold have enough contact surface.
\end{itemize}
\end{promptdoc}

\endgroup

For the real-robot experiments on the Stretch 4 mobile manipulator (\cref{sec:real-world-eval}), the agent receives a different system prompt that describes the robot's kinematics and coordinate frames, the 2D interface with head and wrist camera feeds and a virtual target gripper overlay, and the three waypoint types (navigation, pose, and gripper).
Internal project names and file paths are redacted.

\begingroup
\setlist[itemize]{leftmargin=1.5em, itemsep=1pt, topsep=2pt, parsep=0pt}

\begin{promptdoc}{Robot Control Guide (Stretch 4)}
You operate a Stretch4 robot through the MCP server to complete navigation and manipulation tasks.

\smallskip\noindent\textbf{Robot Information.}
The robot has
\begin{itemize}
\item An omnidirectional mobile base: it can translate in any horizontal direction and rotate in place.
\item A vertical torso fixed to the base; the arm rides up and down along it (robot z-axis).
\item A horizontal arm that telescopes (extends or retracts) along its own axis (robot x-axis).
\item A parallel-jaw gripper with jaw aperture 0--8 cm on a 3-DOF wrist at the end of the arm.
\end{itemize}

\smallskip\noindent\textbf{Axes.}
We have two sets of axes that are handy for different tools:

\textbf{Robot-centric axes}:
\begin{itemize}
\item x: forward --- where the arm points and extends.
\item y: left --- 90 degrees counter-clockwise from x.
\item z: up --- along the vertical torso.
\end{itemize}
The robot axes are attached to the base and rotate with the base.
They are handy for reasoning about navigation, as well as coarse alignment via base rotation and translation.

\textbf{Wrist-centric axes}:
\begin{itemize}
\item x: forward --- the approach axis, where the fingers point.
\item y: left --- the opening axis, the line the jaws open and close along.
\item z: up --- perpendicular to both x and y, through the jaw center.
\end{itemize}
The wrist axes ride and rotate with the gripper.
For example, an \texttt{ee\_wrist\_centric\_rotate} about the x-axis will change both the y- and z-axes.
They match the robot axes only when the gripper is level and pointing forward.
They are handy for reasoning about fine adjustments via \texttt{ee\_wrist\_centric\_translate} / \texttt{ee\_wrist\_centric\_rotate}.

\smallskip\noindent\textbf{Control Interface.}
You control the robot to perform tasks by observing through its cameras and commanding it through a sequence of waypoints using the MCP tools, similar to 3D design software or a 3D game.

\textbf{Observation}:
\begin{itemize}
\item The observation is a collection of images and proprio status.
\item The images are from a head camera (headview) and a wrist camera (wristview).
\item The head camera:
  \begin{itemize}
  \item Fixed relative to the robot base.
  \item Moves and rotates with the base.
  \item Good for scene understanding, navigation, long-distance reaching.
  \item The head view is about 65 degrees wide. When scanning for something, rotate at most 45 to 60 degrees per step so consecutive views overlap.
  \end{itemize}
\item The wrist camera:
  \begin{itemize}
  \item Mounted next to the gripper.
  \item Moves with the base as well as the arm.
  \item Good for fine adjustment and grasp analysis.
  \end{itemize}
\item Both views also contain a virtual \textbf{target gripper}:
  \begin{itemize}
  \item In headview, it is a red gripper overlay.
  \item In wristview, it is a red \texttt{|-{}-{}-{}-.-{}-{}-{}-|} glyph.
  \item Every tool that moves the target gripper will return views with an updated rendering of the virtual gripper, reflecting where the real gripper will go without moving the real robot.
  \item It is the main medium for reasoning and planning your waypoints before executing them.
  \end{itemize}
\end{itemize}

\textbf{Waypoint}:

There are 3 types of waypoints:
\begin{itemize}
\item nav:
  \begin{itemize}
  \item Issued by one or more calls to \texttt{nav\_translate} and \texttt{nav\_rotate}.
  \item Moves the omni base only and does not change the pose of the end-effector relative to the base.
  \item Good for navigation, long-distance movement, or searching for objects of interest.
  \item Be careful to avoid collisions when navigating.
  \end{itemize}
\item pose:
  \begin{itemize}
  \item Issued by one or more calls to \texttt{ee\_\{...\}} tools.
  \item Moves the end-effector to the desired pose via a whole-body controller that moves both the base and end-effector.
  \item Good for object-centric movement (\texttt{ee\_teleport\_via\_click}), or close-up manipulation (\texttt{ee\_robot\_centric\_translate}, \texttt{ee\_wrist\_centric\_translate}, \texttt{ee\_wrist\_centric\_rotate}).
  \end{itemize}
\item gripper:
  \begin{itemize}
  \item Issued by \texttt{gripper\_set}, setting the gripper to open or closed.
  \end{itemize}
\end{itemize}
More about waypoints:
\begin{itemize}
\item For simplicity, a waypoint can only be in one of the three modes.
\item Within the same mode, multiple tool calls can be made to achieve a desired target.
\end{itemize}

\smallskip\noindent\textbf{Workflow.}
\begin{itemize}
\item \texttt{observe}: the initial call to understand the scene and come up with a plan of waypoints.
\item Repeat until the task is complete, one waypoint at a time:
  \begin{itemize}
  \item Adjust the \textbf{target} in a mini-loop: call a tool, inspect the returned views, and call again until the virtual gripper sits where you want it. No real motion happens in this loop.
  \item \texttt{execute\_waypoint}: a controller will bring the real robot to the target pose via linear interpolation while obeying physical constraints.
  \item Check-and-adjust: after the waypoint is reached, analyze its real outcome and adjust the plan for the subsequent waypoints if needed.
  \end{itemize}
\item \texttt{end\_episode} when you believe the task is done, or has become unachievable.
\end{itemize}
For example, to pick up a cube:
\begin{itemize}
\item Locate the cube in the headview.
\item Move close to it.
\item Descend and re-orient the gripper until the cube sits between the fingers.
\item Close the gripper.
\item Lift.
\end{itemize}
Each step is one waypoint that you set and then execute before moving on to the next.

\smallskip\noindent\textbf{Tips.}
\begin{itemize}
\item \texttt{hover} returns the offset of the hovered point (when available) using depth, but the depth is noisy and the depth from the wrist and head cameras may not agree, especially when the object is far away or off-center. Therefore, always trust the images and do not over-rely on the offset information from \texttt{hover}.
\item A common failure is grasping short: closing while the object is still in front of the finger pads, or barely at their tips. Make sure that the fingers and the object have \textbf{enough} contact area. Advance until the intended contact sits deep between the pads before closing. With open jaws, slightly overshooting is usually safer than undershooting.
\end{itemize}
\end{promptdoc}

\endgroup

\subsection{Simulation Task Prompts}
\label{sec:appendix-task-prompts}

Each episode's prompt has two parts: a task instruction embedded in a fixed user-message template, and a task-specific tips file appended to the agent's system prompt.
For the three LIBERO-Goal tasks, Square, and T-block, we provide both \emph{default} and \emph{detailed} prompts; the detailed versions contain example lists of waypoints.
Rainbow has only a default prompt.

\begingroup
\setlist[itemize]{leftmargin=1.5em, itemsep=1pt, topsep=2pt, parsep=0pt}
\setlist[enumerate]{leftmargin=1.5em, itemsep=1pt, topsep=2pt, parsep=0pt}

\smallskip\noindent\textbf{Turn on stove.}\\
Task instruction: ``Turn on the stove.''

\begin{promptdoc}{Turn on stove: default prompt}
The goal is to turn on the stove, which requires rotating the black control knob on the back of the stove by 90 degrees counter-clockwise viewed from the top.

Think carefully about what the task actually requires and about the tools available to you.
Plan an approach that reaches the goal with precision and efficiency.
\end{promptdoc}

\begin{promptdoc}{Turn on stove: detailed prompt}
To turn on the stove, we need to rotate the black control knob on the back of the stove by 90 degrees counter-clockwise.

Example waypoints:
\begin{enumerate}
\item Locate the black control knob on the back of the stove and use \texttt{gripper\_teleport\_via\_click} to teleport the gripper on top of the knob.
\item Lower the gripper towards the knob, and adjust the gripper if needed so that the stem of the knob is roughly in the middle of the fingers.
\item Close the gripper to hold the stem of the knob.
\item Rotate the gripper 90 degrees counter-clockwise to turn the stove on.
\end{enumerate}
\end{promptdoc}

\smallskip\noindent\textbf{Open drawer.}\\
Task instruction: ``Open the middle (2nd from the top) drawer of the cabinet.''

\begin{promptdoc}{Open drawer: default prompt}
The goal is to open the middle (2nd from the top) drawer of the cabinet by pulling it out by its handle.

Think carefully about what the task actually requires and about the tools available to you.
Plan an approach that reaches the goal with precision and efficiency.
\end{promptdoc}

\begin{promptdoc}{Open drawer: detailed prompt}
The trick to opening a drawer is to \textbf{first} rotate the gripper so that the approach axis will point towards the handle when we use \texttt{gripper\_teleport\_via\_click}.
Then clicking the handle lands the gripper $\sim$0.07~m directly in front of it, already aligned to drive straight in.

Example waypoints:
\begin{enumerate}
\item Rotate the gripper so its approach axis points roughly horizontally at the handle.
\item Use \texttt{gripper\_teleport\_via\_click} on the \textbf{handle} of the \textbf{correct} drawer.
\item Move forward along the approach axis until the handle is between the open fingers.
Check in the wrist view that the handle is centered.
Use \texttt{gripper\_translate} or \texttt{gripper\_rotate} to center the handle between the jaws if it is off.
\item Close the gripper.
\item Pull straight back along the approach axis, keeping the gripper level, until the drawer is fully open (may take more than one pull).
\end{enumerate}
\end{promptdoc}

\smallskip\noindent\textbf{Put bowl on plate.}\\
Task instruction: ``Put the bowl on the plate.''

\begin{promptdoc}{Put bowl on plate: default prompt}
The goal is to pick up the bowl and place it on the plate.
Success requires the bowl to end up resting at the \textbf{exact center} of the plate.

Think carefully about what the task actually requires and about the tools available to you.
Plan an approach that reaches the goal with precision and efficiency.
\end{promptdoc}

\begin{promptdoc}{Put bowl on plate: detailed prompt}
\noindent\textbf{How to pick up a bowl.}
To pick up a bowl, we need to pinch it by its side wall with a straight top-down descent.
Here we give example waypoints assuming we pick it from the right ($+Y$ direction).

Example waypoints:
\begin{enumerate}
\item Locate the bowl, and click on the rightmost point on the rim (pinch point).
Trick: The click point \textbf{must} land on the bowl not on the table; after \texttt{gripper\_teleport\_via\_click}, read the salient point's $Y$ from the output; nudge the click along the rim and re-check until $Y$ stops increasing while the click point still lands on the bowl not on the table.
\item Descend the gripper low for a firm grasp.
\item Close the gripper.
\item Lift the gripper.
After a lift, judge grasp success from the point cloud (frontview could be helpful), not the wrist view.
\end{enumerate}

\smallskip\noindent\textbf{How to place onto a plate.}
This task requires the held object to be placed \textbf{exactly at the center} to claim a success.
When holding a bowl, the held object's center is offset from the gripper, so we need to account for that when deciding the target location for the gripper.

Example waypoints:
\begin{enumerate}
\item \texttt{gripper\_teleport\_via\_click} on the desired location of the plate so that the held object will roughly appear above the center of the plate.
E.g. holding the bowl by its right ($+Y$) rim, click on the right ($+Y$) half of the plate.
Adjust the height after clicking to avoid collision.
\item Observe the outcome of the first waypoint and adjust the gripper if needed.
Do NOT solely trust the readout from \texttt{hover}.
Check the visual from the camera and 3D point cloud carefully as well.
\item Descend until the object rests on the plate (contact stops the descent), with a final small $XY$ nudge if needed.
\item Open the gripper to release the held object.
\end{enumerate}
\end{promptdoc}

\smallskip\noindent\textbf{Square.}\\
Task instruction: ``Pick up the square nut and place it onto the square peg.''

% Source: ../../prompts/square_min.md (relative to the paper root).
\begin{promptdoc}{Square: default prompt}
The goal is to pick up the square nut and place it onto the square peg (not the round peg).
Success requires the square peg inside the nut's hole and the nut slid all the way down so it rests near the table.

Think carefully about what the task actually requires and about the tools available to you.
Plan an approach that reaches the goal with precision and efficiency.
\end{promptdoc}

% Source: ../../prompts/square_detailed.md (relative to the paper root).
\begin{promptdoc}{Square: detailed prompt}
\noindent\textbf{How to pick up the square nut:}

Pinch the nut's \textbf{projecting tab} from above, leaving the square hole clear for the peg.

Example waypoints:
\begin{enumerate}
\item Use \texttt{gripper\_teleport\_via\_click} on the exposed tab, where both fingers can descend beside it without hitting the ring.
\item With the gripper pointing down, align the jaws \textbf{across the tab's narrow width}.
Center the tab between them and descend deep enough to straddle its thickness.
\item Close the gripper.
\item Lift.
Verify in the point cloud or 3rd-person view that the whole nut rises with the gripper.
\end{enumerate}

\smallskip\noindent\textbf{How to place onto the square peg:}

Center the \textbf{hole} over the square peg, accounting for its offset from the gripper.
Rotating the held nut changes this offset.

Example waypoints:
\begin{enumerate}
\item Lift until the nut's bottom clears the peg before moving sideways.
Keep it level and rotate until the hole's sides are parallel to the peg's sides.
\item Click the peg's top face, then adjust for the gripper-to-hole offset and height before executing.
Check the actual hole-over-peg alignment in the cameras and point cloud.
\item Descend in small steps, watching the peg enter the hole.
If the nut catches or tilts, lift clear and correct the alignment.
\item Keep holding the tab as the nut slides down.
Once it rests near the table with the peg passing through its hole, open the gripper.
\end{enumerate}
\end{promptdoc}

\smallskip\noindent\textbf{Rainbow.}\\
Task instruction: ``Rearrange the blocks so that it forms a rainbow.
Be creative.''

\begin{promptdoc}{Rainbow: default prompt}
Note that this is a creativity task with no built-in reward/termination detector.
Call \texttt{end\_episode} when you are done.

You may accidentally knock over some blocks while arranging others, therefore you should double check the final arrangement before calling \texttt{end\_episode}.
\end{promptdoc}

\smallskip\noindent\textbf{T-block.}\\
Task instruction: ``Build a structure exactly two blocks tall: the blue block on the bottom, and both green blocks resting directly on top of it, side by side at the same height.
Neither green block may sit on the other or on the table.''

\begin{promptdoc}{T-block: default prompt}
The scene has three cubes of the \textbf{same size}: one blue and two green.

Goal: build a structure that is exactly \textbf{two cubes tall}:
\begin{itemize}
\item Bottom layer: the blue cube, resting on the table.
\item Top layer: \textbf{both} green cubes, sitting side by side on top of the blue cube.
\end{itemize}

For the result to count as correct, \textbf{all} of the following must hold:
\begin{itemize}
\item Both green cubes rest \textbf{directly} on the blue cube --- each green cube touches the blue cube's top face.
\item The two green cubes are at the \textbf{same} height (same level), side by side.
\item The final structure is \textbf{two cubes tall}, never three.
\item The final structure is stable: each green cube has enough contact/support area on the blue cube that it will not tip or fall.
\end{itemize}

Each cube is the same size, so two greens side by side are twice as wide as the blue:

\smallskip\noindent
\texttt{[green][green]~~~<- top layer: both greens, same level, both on the blue}\\
\texttt{\phantom{xxx}[blue~]~~~~~~~<- bottom layer: blue on the table (same size as one green)}\\
\texttt{==============~~~~~~table}
\smallskip

Incorrect (do NOT do this):
\begin{itemize}
\item A three-tall tower: one green on the blue, and the second green stacked on top of the first green.
\item Either green cube resting on the table, or on the other green cube, instead of directly on the blue cube.
\item The two green cubes ending up at different heights.
\end{itemize}

Tips:
Note that because each green cube is exactly as wide as the blue cube's top face, two greens placed side by side in the blue's default orientation cannot both get enough support area --- at least one will overhang too far and tip off.
Reason about how to reorient the blue cube so that both greens gain enough contact area to rest stably, side by side, at the same level.
\end{promptdoc}

% Source: ../../prompts/t_block_detailed.md (relative to the paper root).
\begin{promptdoc}{T-block: detailed prompt}
The scene has three cubes of the \textbf{same size}: one blue and two green.

Goal: build a structure that is exactly \textbf{two cubes tall}:
\begin{itemize}
\item Bottom layer: the blue cube, resting on the table.
\item Top layer: \textbf{both} green cubes, sitting side by side on top of the blue cube.
\end{itemize}

For the result to count as correct, \textbf{all} of the following must hold:
\begin{itemize}
\item Both green cubes rest \textbf{directly} on the blue cube --- each green cube touches the blue cube's top face.
\item The two green cubes are at the \textbf{same} height (same level), side by side.
\item The final structure is \textbf{two cubes tall}, never three.
\item The final structure is stable: each green cube has enough contact/support area on the blue cube that it will not tip or fall.
\end{itemize}

Each cube is the same size, so two greens side by side are twice as wide as the blue:

\smallskip\noindent
\texttt{[green][green]~~~<- top layer: both greens, same level, both on the blue}\\
\texttt{\phantom{xxx}[blue~]~~~~~~~<- bottom layer: blue on the table (same size as one green)}\\
\texttt{==============~~~~~~table}
\smallskip

Incorrect (do NOT do this):
\begin{itemize}
\item A three-cube-tall tower: one green on the blue, and the second green stacked on top of the first green.
\item Either green cube resting on the table, or on the other green cube, instead of directly on the blue cube.
\item The two green cubes ending up at different heights.
\end{itemize}

\smallskip\noindent\textbf{How to prepare the blue base:}

\begin{itemize}
\item Rotate blue \textbf{45 degrees around the z-axis} so both green cubes can keep their centers over blue's top face.
\end{itemize}

Example waypoints:
\begin{enumerate}
\item Click blue's top-face center with \texttt{gripper\_teleport\_via\_click}, descend for a deep grasp, then close.
\item Lift, rotate about 45 degrees, and lower it flat onto the table.
\item Open and lift clear.
\end{enumerate}

\smallskip\noindent\textbf{How to place the green cubes:}

\begin{itemize}
\item Viewed from above, the greens should sit side by side on blue: one on the left and one on the right, with a minimal gap.
\item Use \texttt{hover} to sample points along the \textbf{front-to-back diagonal of the blue cube's top face}, including its two opposite corners and midpoint.
\item This line divides the support area between the greens: place one green on each side.
\end{itemize}

Example waypoints:
\begin{enumerate}
\item Grasp the left green deeply and centrally; lift it level.
Position it on the left half of blue, with its \textbf{right side parallel and as close as possible to the diagonal without crossing it}.
Center that side along the diagonal.
\item Lower onto blue, release, and lift straight up.
Verify the green's top stays one cube height above blue's; if it tips, retry closer to the diagonal.
\item Grasp the right green from \textbf{front and back}, with the jaws opening parallel to the diagonal, so neither finger hits the left green during placement.
\item Keep the right green's bottom above the left green's top while moving over blue's right half.
Align its \textbf{left side parallel and as close as possible to the diagonal without crossing it}, centered along the line.
Lower vertically, leaving a minimal gap between the greens.
\item Release and lift clear.
Verify both greens remain stable at the same height on blue.
\end{enumerate}
\end{promptdoc}

\endgroup

\subsection{Real-World Task Prompts}
\label{sec:appendix-real-task-prompts}

Task prompts for the real-world experiments in \cref{sec:real-world-eval}.

\begingroup
\setlist[itemize]{leftmargin=1.5em, itemsep=1pt, topsep=2pt, parsep=0pt}

% Source: https://github.com/hengyuan-hu/via-dev/blob/main2d/sphinx/prompts/light.md
% Git blob: a1c1c1b4b6ecdf4bc907cbb102cc85fafac46270
\begin{promptdoc}{Lamp}
Task: Turn on the lamp.
\begin{itemize}
\item Look for a lamp on top of a cabinet in this room.
\item Turn it on by pulling its pull chain.
\item Then release the chain and move away.
\end{itemize}
\end{promptdoc}

% Source: https://github.com/hengyuan-hu/via-dev/blob/main2d/sphinx/prompts/drawer.md
% Git blob: 6ff28ef79b6ac59b61e2ffb416fb2b55f393abc0
\begin{promptdoc}{Shelf}
Task:
\begin{itemize}
\item There is a cabinet with two drawers and a lamp on top.
\item Open the cabinet's top drawer.
\item Find a toy lion on the nearby bookshelf and put it inside the top drawer.
\end{itemize}
\end{promptdoc}

% Source: https://github.com/hengyuan-hu/via-dev/blob/main2d/sphinx/prompts/fold.md
% Git blob: a9dc799f349a54200a278a57274b5812f47ca12c
\begin{promptdoc}{Towel}
Task: Fold the towel on the coffee table along its diagonal.
\end{promptdoc}

% Source: https://github.com/hengyuan-hu/via-dev/blob/main2d/sphinx/prompts/wipe.md
% Git blob: a26c1368ecfbb245668d1afe2f8fb9b6eb1dbb6e
\begin{promptdoc}{Wipe}
Task: Wipe the coffee table.
\begin{itemize}
\item Pull a wipe from the CLOROX can.
\item Wipe out the black from the table.
\end{itemize}
\end{promptdoc}

% Source: https://github.com/hengyuan-hu/via-dev/blob/main2d/sphinx/prompts/basket.md
% Git blob: b96e26280142ea5452685fa28ed08a319f61feae
\begin{promptdoc}{Basket}
Task: Fully open the lid of the tall brown basket.
\end{promptdoc}

\endgroup

\end{document}